\documentclass[10pt,twocolumn,letterpaper]{article}

\usepackage{cvpr}              

\usepackage{color}
\usepackage{colortbl}
\usepackage{algorithm}
\usepackage[noend]{algpseudocode} 
\usepackage{amsmath}
\usepackage{booktabs}
\usepackage{multirow}

\definecolor{tableBlue}{rgb}{0.424,0.557,0.749}
\definecolor{lightblue}{HTML}{dfebf7}
\definecolor{verylightgray}{HTML}{EDEDED}
\definecolor{persiangreen}{rgb}{0.0, 0.65, 0.58}
\definecolor{cadmiumred}{rgb}{0.89, 0.0, 0.13}
\newcommand{\tableLineColorGray}{\rowcolor{verylightgray}}

\usepackage{newfloat}
\usepackage{listings}
\usepackage[most]{tcolorbox} 
\usepackage{adjustbox}
\usepackage{amsfonts}
\usepackage{threeparttable}

\definecolor{tableBlue}{rgb}{0.424,0.557,0.749}
\definecolor{lightblue}{HTML}{dfebf7}
\newcommand{\TableLineColor}{\rowcolor{lightblue}}
\definecolor{persiangreen}{rgb}{0.0, 0.65, 0.58}
\definecolor{cadmiumred}{rgb}{0.89, 0.0, 0.13}
\definecolor{increaseGreen}{rgb}{0.51,0.702,0.4}
\definecolor{decreaseRed}{rgb}{0.647,0.337,0.318}

\usepackage{amsthm}
\usepackage{booktabs}

\usepackage{pifont}
\usepackage{makecell}
\usepackage{titletoc}
\titlecontents{section}[3.8em]
  {}
  {\hyperlink{section.\thecontentslabel}{\contentslabel{1.8em}}}  
  {\hspace*{-2.em}}
  {\titlerule*[0.5pc]{.}\contentspage}

\titlecontents{subsection}[6.5em]
  {}
  {\hyperlink{subsection.\thecontentslabel}{\contentslabel{2.5em}}}
  {\hspace*{-3.em}}
  {\titlerule*[0.5pc]{.}\contentspage}

\tcbset{
    promptboxstyle/.style={
        colframe=blue!75!black,    
        colback=blue!5,           
        coltitle=white,           
        fonttitle=\bfseries,      
        sharp corners,            
        width=\linewidth,                
        boxrule=1pt,              
        arc=4mm,                  
        drop shadow,              
        enhanced,         
        breakable
    }
}

\tcbset{
    exampleboxstyle/.style={
        colframe=gray!75!black,    
        colback=blue!5,           
        coltitle=white,           
        fonttitle=\bfseries,      
        sharp corners,            
        width=14cm,               
        boxrule=1pt,              
        arc=4mm,                  
        drop shadow,             
        enhanced,                
    }
}

\definecolor{cvprblue}{rgb}{0.21,0.49,0.74}
\usepackage[pagebackref,breaklinks,colorlinks,allcolors=cvprblue]{hyperref}

\def\paperID{16439} 
\def\confName{CVPR}
\def\confYear{2026}

\title{Towards Efficient Medical Reasoning with Minimal Fine-Tuning Data}

\author{
    Xinlin Zhuang\textsuperscript{1,2} \quad
    Feilong Tang\textsuperscript{2,3} \quad
    Haolin Yang\textsuperscript{2} \quad
    Xiwei Liu\textsuperscript{2} \quad
    Ming Hu\textsuperscript{2,3,4} \quad
    Huifa Li\textsuperscript{2} \quad  \\
    Haochen Xue\textsuperscript{2} \quad
    Junjun He\textsuperscript{4} \quad
    Zongyuan Ge\textsuperscript{3} \quad
    Yichen Li\textsuperscript{2} \quad
    Ying Qian\textsuperscript{1}\thanks{Corresponding authors.} \quad
    Imran Razzak\textsuperscript{2}\footnotemark[1] \\[6pt] 
    \textsuperscript{1}East China Normal University \quad
    \textsuperscript{2}MBZUAI \quad
    \textsuperscript{3}Monash University \quad
    \textsuperscript{4}Shanghai AI Laboratory \\ [4pt]
    {\tt\small xinlinzhuang@stu.ecnu.edu.cn, yqian@cs.ecnu.edu.cn, imran.razzak@mbzuai.ac.ae}
}

\begin{document}
\maketitle
\begin{abstract}
Supervised Fine-Tuning (SFT) of the language backbone plays a pivotal role in adapting Vision-Language Models (VLMs) to specialized domains such as medical reasoning. 
However, existing SFT practices often rely on unfiltered textual datasets that contain redundant and low-quality samples, leading to substantial computational costs and suboptimal performance in complex clinical scenarios.
Although existing methods attempt to alleviate this problem by selecting data based on sample difficulty, defined by \textit{knowledge} and \textit{reasoning} complexity, they overlook each sample's optimization utility reflected in its gradient. 
Interestingly, we find that gradient-based influence alone favors easy-to-optimize samples that cause large parameter shifts but lack deep reasoning chains, while difficulty alone selects noisy or overly complex textual cases that fail to guide stable optimization. 
Based on this observation, we propose a data selection strategy, \textit{\textbf{D}ifficulty-\textbf{I}nfluence \textbf{Q}uadrant} \textbf{(DIQ)}, which prioritizes samples in the "high-difficulty-high-influence" quadrant to balance complex clinical reasoning with substantial gradient influence.
This enables efficient medical reasoning for VLMs with minimal fine-tuning data.
Furthermore, Human and LLM-as-a-judge evaluations show that DIQ-selected subsets demonstrate higher data quality and generate clinical reasoning that is more aligned with expert practices in \textit{differential diagnosis}, \textit{safety check}, and \textit{evidence citation}, as DIQ emphasizes samples that foster expert-like reasoning patterns. 
Extensive experiments on medical reasoning benchmarks demonstrate that DIQ enables VLM backbones fine-tuned on only \textbf{1\%} of selected data to match full-dataset performance, while using \textbf{10\%} consistently outperforms baseline methods, highlighting the superiority of principled data selection over brute-force scaling. 
The code is available at \url{https://github.com/mihara-bot/DIQ}.
\end{abstract}    
\section{Introduction}
\label{sec:intro}
Large Language Models (LLMs)~\cite{deepseek,touvron2023llama}, which serve as the core reasoning engines for modern Vision-Language Models (VLMs), have achieved notable success in reasoning-intensive tasks such as mathematics and programming~\cite{deepseek,qwq}.
Inspired by this progress, the medical community is exploring these foundation models for high-stakes scenarios such as clinical diagnosis and treatment planning, and the prospect of language backbones that emulates clinician cognitive processes is transforming the landscape of healthcare services, especially in multimodal applications~\cite{mcduff2025towardsddx,tordjman2025comparative,wu2025knowledge}.
However, current models still face significant challenges in synthesizing incomplete and ambiguous clinical information (often derived from complex visual findings) into reliable decisions~\cite{medfound, disentangling}.

\begin{figure}[t]
    \centering
    \setlength{\abovecaptionskip}{0cm}
    \includegraphics[width=\linewidth]{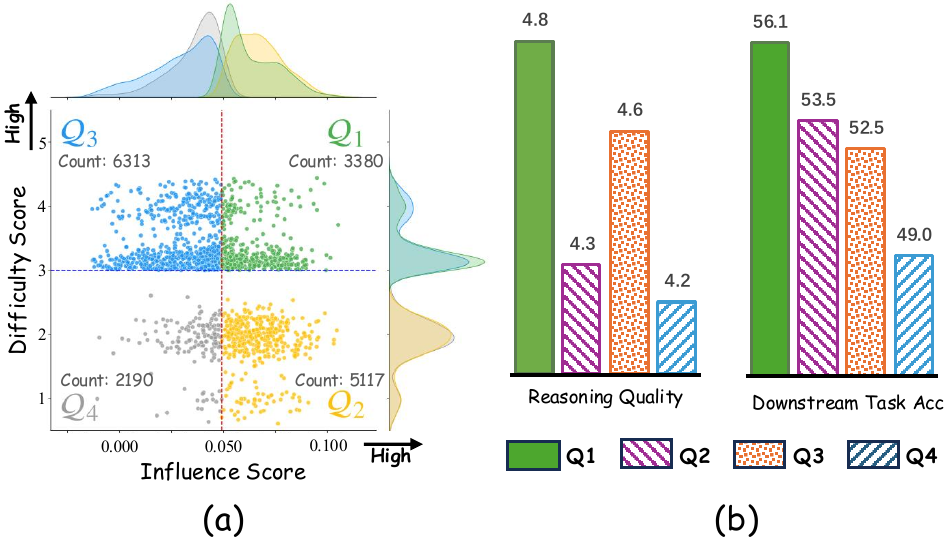}
    \caption{\textbf{(a)} The FineMed dataset distributed by \textbf{difficulty} and \textbf{dot influence} scores, with data points colored by quadrant. \textbf{(b)} For each quadrant, the bar shows the intrinsic reasoning quality of the data and the resulting downstream performance of a Qwen3-8B model fine-tuned on the corresponding data subset.}
    \label{fig:intro}
\end{figure}

Supervised Fine-Tuning (SFT) is a prevailing practice for adapting LLM backbonesto medical domains.
Following the assumption that larger datasets improve performance, previous works explored scaling up medical datasets with tens of thousands of Chain-of-Thought (CoT) examples~\cite{ultramedical,reasonmed}. 
For instance, ReasonMed employs multi-agent verification to curate 370k instances~\cite{reasonmed}. However, reliance on unfiltered data with redundant and low-value samples incurs substantial computational costs and inefficient training. Early data selection efforts have relied on coarse, \textbf{difficulty}-based heuristics, such as removing easy samples or retaining only those unsolved after multiple trials~\cite{huatuogpto1,medreason}. While more recent studies refine this by decoupling difficulty into \textit{Knowledge} and \textit{Reasoning} complexity~\cite{disentangling, m1}, these approaches are still limited. Critically, they remain one-dimensional, evaluating samples in isolation while overlooking their \textbf{optimization utility}~\cite{if,tracin} as reflected by the training gradient. Consequently, the precise impact of difficult samples in training, encompassing both their benefits and drawbacks, is not yet fully understood.
\par
To investigate the interplay between sample difficulty and influence, we conducted a pilot experiment on the FineMed dataset \cite{finemedlm}. 
We partitioned the data into four quadrants based on these two metrics (Fig.~\ref{fig:intro} (a)) and then evaluated each quadrant for model performance (via fine-tuning on them) and reasoning quality, as assessed by Gemini-2.5-Pro \cite{gemini25}, as detailed in Fig.~\ref{fig:intro} (b) and App.~\ref{app:pilot}.
The results reveal a critical tension. 
Models fine-tuned on high-influence, low-difficulty $\mathcal{Q}_2$ data consistently outperformed those trained on low-influence, high-difficulty data $\mathcal{Q}_3$, even though the samples in $\mathcal{Q}_2$ are of lower reasoning quality.
We found this occurs because $\mathcal{Q}_2$ samples, while easy to optimize, possess shallower reasoning chains. 
Conversely, $\mathcal{Q}_3$ samples are reasoning-intensive but provide weak gradient signals, leading to unstable training and lower downstream performance.
These findings expose an unavoidable flaw in one-dimensional selection: prioritizing influence alone favors simplistic samples, while prioritizing difficulty alone selects for noisy, hard-to-learn cases.
\par
In this paper, we propose a simple yet effective data selection strategy for textual medical reasoning, \textit{\textbf{D}ifficulty-\textbf{I}nfluence \textbf{Q}uadrant} \textbf{(DIQ)}, which prioritizes samples in the \textit{high-difficulty-high-influence} quadrant to balance complex clinical reasoning with substantial influence score, to enable efficient medical reasoning with minimal data. 
Specifically, the \textit{Difficulty} for each medical sample is obtained from a classifier constructed by fine-tuning BiomedBERT \cite{biomedbert} on medical questions collected from multiple medical QA datasets and assesses the complexity of \textit{knowledge} and \textit{reasoning} on a 5-point Likert scale. 
Meanwhile, the \textit{Influence} quantifies the expected impact of each instance on model improvement, efficiently calculated by aggregating first-order gradient dot products between training and validation samples across tasks and epochs. 
This lightweight approach enables scalable influence approximation while avoiding the high computational cost of traditional methods.
Based on these two scores, our selection strategy proceeds in the order of $\mathcal{Q}_1$, $\mathcal{Q}_2$, $\mathcal{Q}_3$ and $\mathcal{Q}_4$ as defined in Fig.~\ref{fig:intro} (a), to ensure that the selected subset maximally balances the complexity of reasoning and the utility of training. 
\par
To assess the effectiveness of DIQ, we perform human and LLM-as-a-judge evaluations on DIQ-selected subsets, showing that they exhibit higher data quality and generate clinical reasoning closely aligned with expert practices in \textit{differential diagnosis}, \textit{safety check}, and \textit{evidence citation} (Tab.~\ref{tab:clinical_value}). 
These results demonstrate that DIQ selects samples that foster expert-like reasoning patterns. 
Extensive experiments on medical reasoning benchmarks demonstrate that DIQ enables models fine-tuned on \textbf{only 1\%} of selected data to match full-dataset performance, while \textbf{10\%} consistently outperforms baseline methods, highlighting the superiority of principled data selection over brute-force scaling.
Our contributions can be summarized as follows:
\begin{itemize}
    \item We propose DIQ, a principled data selection framework for medical SFT that jointly measures sample difficulty and optimization influence, enabling more efficient medical reasoning with fewer fine-tuning data, thereby building robust language foundations for medical VLMs.
    \item We conduct a comprehensive evaluation showing that DIQ-selected subsets improve data quality and the alignment of clinical reasoning with expert practices.
    \item We demonstrate empirically that training on only 1\% of DIQ-selected data matches the performance of the full dataset, while 10\% consistently surpasses it.
\end{itemize}
\section{Related Work}
\label{sec:related_work}

\paragraph{Medical Reasoning Dataset Construction.}
The construction of high-quality medical reasoning datasets has emerged as a central focus, with methodologies divided into two distinct paradigms.
The predominant approach involves creating large-scale datasets through synthetic data generation using powerful foundation models like GPT-4o within multi-agent verification pipelines (e.g., UltraMedical \cite{ultramedical}, ReasonMed \cite{reasonmed}), or by leveraging medical knowledge graphs and multi-stage training procedures to embed clinical validity and complex reasoning pathways (e.g., MedReason \cite{medreason}, FineMedLM-o1 \cite{finemedlm}, HuatuoGPT-o1 \cite{huatuogpto1}). Despite driving recent progress, large synthetic datasets remain costly and labor-intensive to curate consistently. In contrast, a compelling line of research \cite{m1, lima, limo} posits that for base models with extensive domain knowledge, complex reasoning capabilities can be effectively elicited with remarkably few high-quality examples. Inspired by this \textit{less is more} premise, our work investigates data-efficient fine-tuning for medical reasoning.
We demonstrate that strategically selecting a small subset from existing medical SFT datasets can achieve performance comparable or even superior to full-dataset fine-tuning.
\vspace{-1em}
\paragraph{Medical Reasoning Benchmarking.}
Rigorous evaluation methodology is paramount for advancing medical reasoning models.
The assessment landscape has evolved from factual recall to sophisticated clinical analysis.
Foundational benchmarks consist of multiple-choice question-answering datasets such as MedQA \cite{medqa}, MedMCQA \cite{medmcqa}, and the medical portions of MMLU-Pro \cite{mmlupro}, which gauge base medical knowledge.
As state-of-the-art model performance on these knowledge-intensive tasks approaches saturation, the community has recognized the need for more challenging assessments.
Consequently, a new wave of benchmarks has emerged to probe deeper cognitive abilities, including datasets grounded in complex clinical case analysis like MedBullets \cite{medbullets} and MedXpertQA \cite{medxpertqa}, which require synthesizing patient data and forming differential diagnoses.
Expert-level challenges such as HLE \cite{hle} further assess complex, cross-domain reasoning.
Our evaluation suite of nine distinct benchmarks is designed to provide a holistic assessment, examining capabilities from fundamental knowledge recall to sophisticated clinical reasoning.
\begin{figure*}[!t]
  \centering
  \includegraphics[width=1.0\textwidth]{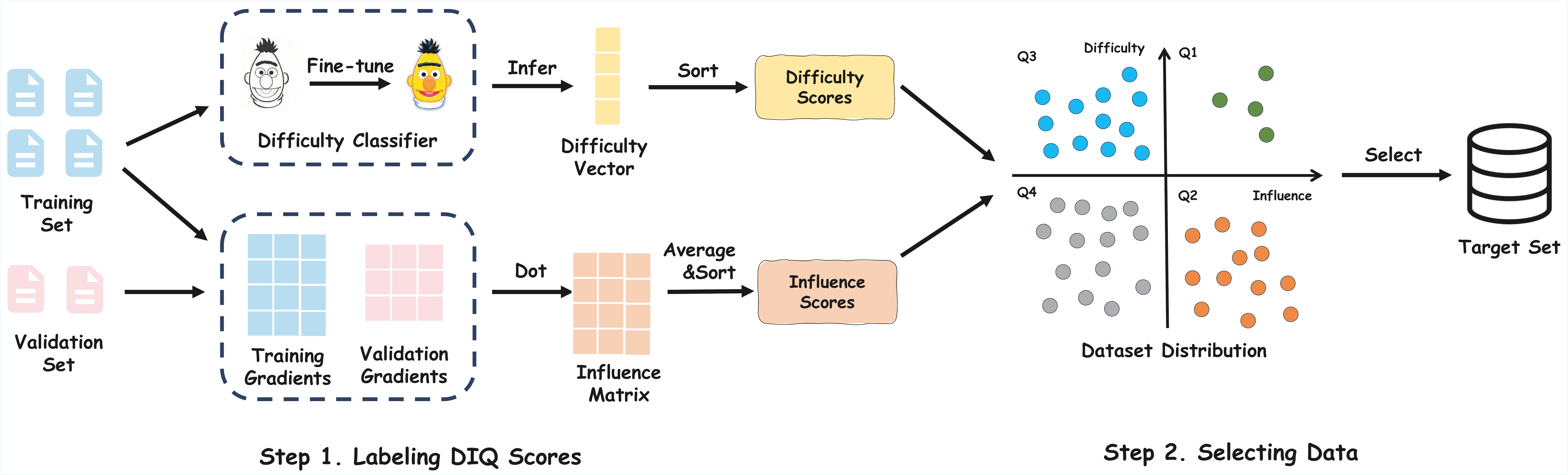}.
  \caption{Overview of the DIQ framework. Each sample is projected to a two-dimensional space using (i) a BiomedBERT classifier to produce a scalar \textbf{difficulty} score (one chosen dimension among \textit{Knowledge}/\textit{Reasoning}/\textit{Overall};  Sec.~\ref{Sec:Difficulty}), and (ii) an \textbf{influence} score, \textit{Dot}, computed as the inner product between the sample gradient and the mean validation gradient (Eq.~\ref{eq:dot}). Using a percentile threshold on difficulty and the median of Dot, the dataset is partitioned into four quadrants. DIQ selects data by priority ($\mathcal{Q}_1 \to \mathcal{Q}_2 \to \mathcal{Q}_3 \to \mathcal{Q}_4$), sorting within each quadrant by Dot (ties by difficulty), until the target retention ratio is reached.}
  \label{fig:workflow}

\end{figure*}

\vspace{-1em}
\section{Method}
\label{sec:method}

\subsection{Overview}
In this section, we propose \textbf{DIQ}, in which each sample receives (i) a \textbf{difficulty} score that reflects intrinsic reasoning complexity (model-agnostic), and (ii) an \textbf{influence} score that measures its expected validation-loss reduction for the current model (model-dependent). We instantiate influence with a simple and scalable metric, \textbf{Dot}, defined via gradient inner products without Hessian terms. 
These two scores span a 2D space where we partition data into four quadrants and select by priority to meet the target data retention ratio $r$.
Fig.~\ref{fig:workflow} illustrates the overview of our DIQ method.

Given a pre-trained LLM with parameters $\boldsymbol{\theta}$ and a medical reasoning dataset $\mathcal{D}=\{z_i\}_{i=1}^N$ in instruction-following format where each instance $z=(q,a)$ contains an input $q$ and a reference answer $a$ with reasoning trace, the target of SFT is to minimize the empirical risk
\begin{equation}
\boldsymbol{\theta}^* = \arg\min_{\boldsymbol{\theta}} \mathbb{E}_{(q,a) \sim \mathcal{D}} [\ell(q, a; \boldsymbol{\theta})],
\end{equation}
where $\ell$ denotes the token-level cross-entropy loss. 
Our goal is to select a subset $\mathcal{S}\subset\mathcal{D}$ with the pre-defined retention ratio $r\in(0,1)$ such that SFT on $\mathcal{S}$ maximizes downstream performance of the model. 

\subsection{Difficulty Estimation}
\label{Sec:Difficulty}

Inspired by previous work \cite{disentangling}, we estimate difficulty along three ordinal dimensions: \textit{Knowledge}, \textit{Reasoning}, and \textit{Overall}, each annotated on a 5-point Likert scale and predicted by a BiomedBERT-based classifier \cite{biomedbert}. 
In DIQ, we use a single dimension as the difficulty scalar. 
Formally, let $D_{\mathrm{K}}(z)$, $D_{\mathrm{R}}(z)$, and $D_{\mathrm{O}}(z)$ denote the calibrated scores on the three dimensions for sample $z$. 
We choose one dimension $\phi\in\{\mathrm{K},\mathrm{R},\mathrm{O}\}$ and define the difficulty score as
\begin{equation}
\label{eq:difficulty}
D(z)\;\triangleq\; D_{\phi}(z), \qquad \phi\in\{\mathrm{K},\mathrm{R},\mathrm{O}\}.
\end{equation}
The high/low split in DIQ is then determined by a percentile threshold $\tau_d$ on $\{D(z)\}$ (e.g., the $p$-th quantile over the training set). 
We report results for different choices of $\phi$ in ablations (i.e., using \textit{Knowledge}, \textit{Reasoning}, or \textit{Overall} individually as $D$) and discuss their impact on selection and downstream performance in Sec. \ref{sec:ablation}.
Further details of annotation  and training process are provided in App.~\ref{app:difficulty_classifier}.

\subsection{Dot-Product Influence}
\label{Sec:Influence}

For a sample $z$, let $g(z;\boldsymbol{\hat\theta})\triangleq\nabla_{\boldsymbol{\theta}}\ell(z;\boldsymbol{\hat\theta})$ be its per-example gradient evaluated at a reference parameter $\hat{\boldsymbol{\theta}}$ (the base model checkpoint before SFT). 
Let $\mathcal{D}_{\text{val}}$ be a small validation set (we randomly select 20 samples from each downstream task as default), which is used only to compute influence scores.
The mean validation gradient is
\begin{equation}
\bar{g}_{\text{val}}(\hat{\boldsymbol{\theta}})\triangleq
\frac{1}{|\mathcal{D}_{\text{val}}|}\sum_{z'\in\mathcal{D}_{\text{val}}} g(z';\hat{\boldsymbol{\theta}}).
\label{eq:valgrad}
\end{equation}
We define the \textbf{Dot influence} of a training sample $z$ by
\begin{equation}
\label{eq:dot}
\begin{aligned}
\mathrm{Dot}(z) &\triangleq g(z;\hat{\boldsymbol{\theta}})^\top \bar{g}_{\text{val}}(\hat{\boldsymbol{\theta}}) \\
&= \frac{1}{|\mathcal{D}_{\text{val}}|}\sum_{z'\in\mathcal{D}_{\text{val}}} g(z;\hat{\boldsymbol{\theta}})^\top g(z';\hat{\boldsymbol{\theta}}).
\end{aligned}
\end{equation}

Intuitively, $\mathrm{Dot}(z)$ measures the alignment between the training gradient of $z$ and the average validation gradient. \
A larger positive value implies a stronger expected decrease of the average validation loss after updating on $z$ while a negative value suggests potential harm.

Consider one gradient descent step at learning rate $\eta$ on training point $z$, updating $\boldsymbol{\theta}^+=\boldsymbol{\theta}-\eta\,g(z;\boldsymbol{\theta})$. For any validation point $z'$, by a first-order Taylor expansion,
\begin{equation}
\label{eq:taylor}
\begin{aligned}
\ell(z';\boldsymbol{\theta}^+)-\ell(z';\boldsymbol{\theta})
&= -\eta\,g(z;\boldsymbol{\theta})^\top g(z';\boldsymbol{\theta}) \\
&\quad +\tfrac{1}{2}\eta^2\,g(z;\boldsymbol{\theta})^\top H_{z'}(\tilde{\boldsymbol{\theta}})\,g(z;\boldsymbol{\theta}),
\end{aligned}
\end{equation}
where $H_{z'}$ is the Hessian matrix of $\ell(z';\cdot)$ and $\tilde{\boldsymbol{\theta}}$ lies on the segment between $\boldsymbol{\theta}$ and $\boldsymbol{\theta}^+$. Averaging Eq.~\eqref{eq:taylor} over every validation sample $z'\in\mathcal{D}_{\text{val}}$ yields
\begin{equation}
\label{eq:avg-decrease}
\begin{aligned}
\Delta \bar{\ell}_{\text{val}}
&= \frac{1}{|\mathcal{D}_{\text{val}}|}\sum_{z' \in \mathcal{D}_{\text{val}}}\big(\ell(z';\boldsymbol{\theta}^+)-\ell(z';\boldsymbol{\theta})\big) \\
&= -\eta\,\mathrm{Dot}(z) + O(\eta^2).
\end{aligned}
\end{equation}
Hence, up to a second-order remainder, ranking samples by $\mathrm{Dot}(\cdot)$ approximates ranking them by the expected one-step decrease in average validation loss. When the learning rate is small and local curvature is bounded, the $O(\eta^2)$ term is dominated by the first-order term. App.~\ref{app:influence_score} provides the full derivation, discusses extensions to mini-batch SGD/momentum and weight decay, and shows that the same ordering arises if one applies a sample-independent preconditioner (e.g., a fixed diagonal scaling).

In practice, we first compute $\bar{g}_{\text{val}}$ in Eq.~\eqref{eq:valgrad} with a single backward pass per validation sample. Then, for each training sample $z$, we compute one backward pass to obtain $g(z)$ and take the inner product in Eq.~\eqref{eq:dot}. 
This yields an $O(|\mathcal{D}_{\text{val}}| + |\mathcal{D}|)$ backward complexity and requires no Hessian or Hessian-vector products. 
Moreover, we apply a Johnson-Lindenstrauss guaranteed random projection to restrict gradient to a lower-dimension subspace (the default dimension is set as 4096), which preserves ranking well and reduces computational cost enormously.

\begin{algorithm}[!t]
\caption{DIQ Algorithm}
\label{alg:diq}
\begin{algorithmic}[1]
\Require Training set $\mathcal{D}$, difficulty threshold $\tau_d$, retention ratio $r$.
\Ensure Selected subset $\mathcal{S}$.
\State $\mathcal{S} \leftarrow \emptyset$,\quad $N_{\text{target}} \leftarrow \lfloor |\mathcal{D}|\cdot r \rfloor$
\State Compute $D(z)$ for all $z\in\mathcal{D}$ using the difficulty classifier.
\State Construct $\mathcal{D}_{\text{val}}$ and compute $\bar{g}_{\text{val}}$ by Eq.~\eqref{eq:valgrad}.
\State For each $z\in\mathcal{D}$, compute $\mathrm{Dot}(z)=\langle g(z),\bar{g}_{\text{val}}\rangle$ by Eq.~\eqref{eq:dot}.
\State $m_{\text{dot}} \leftarrow \text{median}\big(\{\mathrm{Dot}(z)\,|\, z\in\mathcal{D}\}\big)$
\State Partition $\mathcal{D}$ into $\mathcal{Q}_1,\mathcal{Q}_2,\mathcal{Q}_3,\mathcal{Q}_4$ using $\tau_d$ and $m_{\text{dot}}$.
\For{$\mathcal{Q}$ in $(\mathcal{Q}_1,\mathcal{Q}_2,\mathcal{Q}_3,\mathcal{Q}_4)$}
    \If{$|\mathcal{S}|\ge N_{\text{target}}$} \textbf{break} \EndIf
    \State Let $\mathcal{Q}'$ be $\mathcal{Q}$ sorted by $\mathrm{Dot}(z)$ descending (ties by $D(z)$ descending).
    \State $n_{\text{take}} \leftarrow \min\big(|\mathcal{Q}'|,\, N_{\text{target}}-|\mathcal{S}|\big)$
    \State $\mathcal{S} \leftarrow \mathcal{S}\cup \mathcal{Q}'[1:n_{\text{take}}]$
\EndFor
\State \Return $\mathcal{S}$
\end{algorithmic}
\end{algorithm}

\subsection{Quadrant-Based Data Selection}
\label{Sec:Quadrants}
Having computed the two score dimensions $D(z),\,\mathrm{Dot}(z)$, we partition $\mathcal{D}$ into four quadrants using data-adaptive thresholds. 
Let $\tau_d$ denote the difficulty threshold (a percentile of $\{D(z)\}$), and let $m_{\text{dot}}$ denote the median of $\{\mathrm{Dot}(z)\}$, the four quadrants can be defined as 
\begin{align*}
\mathcal{Q}_1 &= \{z\in\mathcal{D}\,|\, D(z)\ge \tau_d \;\land\; \mathrm{Dot}(z)\ge m_{\text{dot}}\} \\
\mathcal{Q}_2 &= \{z\in\mathcal{D}\,|\, D(z)< \tau_d \;\land\; \mathrm{Dot}(z)\ge m_{\text{dot}}\} \\
\mathcal{Q}_3 &= \{z\in\mathcal{D}\,|\, D(z)\ge \tau_d \;\land\; \mathrm{Dot}(z)< m_{\text{dot}}\} \\
\mathcal{Q}_4 &= \{z\in\mathcal{D}\,|\, D(z)< \tau_d \;\land\; \mathrm{Dot}(z)< m_{\text{dot}}\}.
\end{align*}
We prioritize $\mathcal{Q}_1$ and then fill $\mathcal{Q}_2$, $\mathcal{Q}_3$, and $\mathcal{Q}_4$ sequentially until reaching the target retention $r$. Within each quadrant, we sort samples by $\mathrm{Dot}(z)$ in descending order; ties are broken by $D(z)$ (also descending). This strategy balances \textit{intrinsic challenge} and \textit{optimization utility}.
The complete process of DIQ is shown in Alg.~\ref{alg:diq}.
\section{Experiment}


\subsection{Experimental Setup}
\label{Sec:Data}

\begin{table*}[!t]
\centering
\small
\begin{adjustbox}{max width=1.0\textwidth}
\begin{tabular}{llcccccccccccc}
\toprule
\multicolumn{1}{c}{\multirow{2}{*}{\textbf{Model}}} & \multicolumn{1}{c}{\multirow{2}{*}{\textbf{Data}}} & \multicolumn{4}{c}{\textbf{Clinical Standard Tasks}}                                                                                             & \multicolumn{7}{c}{\textbf{Clinical Challenging Tasks}}                                                                                                                                                                                                     &                                      \\ \cmidrule(l){3-14} 
\multicolumn{1}{c}{}                                & \multicolumn{1}{c}{}                               & \multicolumn{1}{l}{\textbf{MedQ}} & \multicolumn{1}{l}{\textbf{MedM}} & \multicolumn{1}{l}{\textbf{MMLU}} & \multicolumn{1}{l}{\textbf{Avg$_S$}} & \multicolumn{1}{l}{\textbf{HLE}} & \multicolumn{1}{l}{\textbf{MeB4}} & \multicolumn{1}{l}{\textbf{MeB5}} & \multicolumn{1}{l}{\textbf{MedX}} & \multicolumn{1}{l}{\textbf{MedG}} & \multicolumn{1}{l}{\textbf{MetM}} & \multicolumn{1}{l}{\textbf{Avg$_C$}} & \multicolumn{1}{l}{\textbf{Avg$_A$}} \\ \midrule
\multicolumn{14}{c}{\textbf{General Non-reasoning Models}}                                                                                                                                                                                                                                                                                                                                                                                                                                                                                                       \\ \midrule
GPT-4.1                                             & \multicolumn{1}{c}{--}                             & 84.29                             & 73.34                             & 82.46                             & 80.03                                & 7.77                             & 71.75                             & 70.13                             & 42.00                             & 64.44                             & 70.79                             & 54.48                                & 63.00                                \\
DeepSeek-V3-0324                                    & \multicolumn{1}{c}{--}                             & 73.76                             & 55.10                             & 62.90                             & 63.92                                & 6.80                             & 73.38                             & 66.56                             & 38.04                             & 59.77                             & 65.84                             & 51.73                                & 55.79                                \\
Gemini-2.5-flash                                    & \multicolumn{1}{c}{--}                             & 90.73                             & 77.34                             & 90.36                             & 86.14                                & 11.65                            & 82.14                             & 76.62                             & 36.82                             & 61.55                             & 77.13                             & 57.65                                & 67.15                                \\ \midrule
\multicolumn{13}{c}{\textbf{General Reasoning Models}}                                                                                                                                                                                                                                                                                                                                                                                                                                                                    &                                      \\ \midrule
DeepSeek-R1-0528                                    & \multicolumn{1}{c}{--}                             & 92.85                             & 76.55                             & 91.28                             & 86.89                                & 13.59                            & 83.44                             & 54.22                             & 38.61                             & 59.88                             & 72.91                             & 53.78                                & 64.81                                \\
QwQ-32B                                             & \multicolumn{1}{c}{--}                             & 75.10                             & 63.45                             & 78.97                             & 72.51                                & 12.62                            & 67.86                             & 59.09                             & 22.65                             & 48.44                             & 63.80                             & 45.74                                & 54.66                                \\
o4-mini-medium                                      & \multicolumn{1}{c}{--}                             & 64.73                             & 61.44                             & 81.08                             & 69.08                                & 13.59                            & 70.78                             & 71.10                             & 40.78                             & 60.46                             & 76.11                             & 55.47                                & 60.01                                \\
Gemini-2.5-pro                                      & \multicolumn{1}{c}{--}                             & 78.00                             & 79.75                             & 85.67                             & 81.14                                & 15.53                            & 84.42                             & 78.57                             & 42.37                             & 62.11                             & 73.05                             & 59.34                                & 66.61                                \\ \midrule
\multicolumn{13}{c}{\textbf{Fine-tuned Medical Reasoning Models}}                                                                                                                                                                                                                                                                                                                                                                                                                                                         &                                      \\ \midrule
Llama3.1-8B-Instruct                                & \multicolumn{1}{c}{--}                             & 53.26                             & 53.15                             & 61.57                             & 55.99                                & 11.65                            & 37.99                             & 36.04                             & 15.63                             & 42.26                             & 37.22                             & 30.13                                & 38.75                                \\ \midrule
\tableLineColorGray \cellcolor{white}               & Full (19k)                                         & 58.68                             & 47.79                             & 57.85                             & 54.77                                & 24.27                            & 44.16                             & 40.91                             & 20.33                             & 43.28                             & 53.68                             & 37.77                                & 43.44                                \\ \cmidrule(l){2-14} 
                                                    & \textit{1\%}                                       &                                   &                                   &                                   &                                      &                                  &                                   &                                   &                                   &                                   &                                   &                                      &                                      \\
                                                    & Random                                             & 54.75                             & 43.99                             & 55.19                             & 51.31                                & 15.86                            & 42.50                             & 37.71                             & 13.63                             & 44.75                             & 46.39                             & \underline{33.47}                                & 39.42                                \\
                                                    & PPL                                                & 52.96                             & 46.66                             & 60.97                             & 53.53                                & 7.77                             & 46.84                             & 41.88                             & 13.18                             & 45.00                             & 35.69                             & 31.73                                & 38.99                                \\
                                                    & Similarity                                         & 50.22                             & 43.90                             & 62.53                             & 52.22                                & 12.62                            & 43.51                             & 40.91                             & 12.78                             & 41.38                             & 37.51                             & 31.45                                & 38.37                                \\
                                                    & LESS                                               & 56.17                             & 50.15                             & 58.58                             & \underline{54.97}                                & 13.59                            & 46.84                             & 48.70                             & 15.35                             & 40.12                             & 35.32                             & 33.32                                & \underline{40.54}                                \\ 
\TableLineColor \cellcolor{white}                   &  \textbf{DIQ (ours)}                                & 56.64                             & 50.16                             & 62.81                             & \textbf{56.54}                       & 13.59                            & 47.40                             & 47.75                             & 14.45                             & 45.86                             & 46.39                             & \textbf{35.91}                       & \textbf{42.78}                       \\ \cmidrule(l){2-14} 
                                                    & \textit{10\%}                                      &                                   &                                   &                                   &                                      &                                  &                                   &                                   &                                   &                                   &                                   &                                      &                                      \\
                                                    & Random                                             & 52.87                             & 47.33                             & 56.75                             & \underline{55.39}                                & 15.53                            & 39.94                             & 30.84                             & 17.51                             & 40.71                             & 35.32                             & 29.98                                & 37.42                                \\
                                                    & PPL                                                & 50.22                             & 47.82                             & 51.61                             & 49.88                                & 16.50                            & 41.13                             & 38.82                             & 13.18                             & 41.65                             & 53.82                             & 34.18                                & 39.42                                \\
                                                    & Similarity                                         & 53.26                             & 48.12                             & 61.00                             & 54.13                                & 18.45                            & 42.86                             & 40.26                             & 16.45                             & 42.78                             & 52.37                             & \underline{35.53}                                & \underline{41.73}                                \\
                                                    & LESS                                               & 54.01                             & 48.99                             & 61.72                             & 54.91                                & 18.45                            & 41.48                             & 39.61                             & 18.57                             & 42.78                             & 42.37                             & 33.88                                & 40.89                                \\
\TableLineColor \cellcolor{white}\multirow{-13}{*}{Huatuo~\cite{huatuogpto1}} &  \textbf{DIQ (ours)}                                & 58.13                             & 53.57                             & 62.63                             & \textbf{58.11}                       & 25.24                            & 44.48                             & 40.40                             & 17.59                             & 43.38                             & 50.91                             & \textbf{37.00}                       & \textbf{44.04}                       \\ \midrule

\tableLineColorGray \cellcolor{white}               &  Full (17k)                                         & 40.22                             & 51.26                             & 51.61                             & 42.52                                & 16.50                            & 46.10                             & 44.48                             & 25.47                             & 39.27                             & 32.19                             & 34.00                                & 38.57                                \\ \cmidrule(l){2-14} 
                                                    & 1\%                                                &                                   &                                   &                                   &                                      &                                  &                                   &                                   &                                   &                                   &                                   &                                      &                                      \\
                                                    & Random                                             & 51.61                             & 48.98                             & 58.68                             & \underline{53.09}                                & 11.65                            & 45.45                             & 42.86                             & 13.59                             & 40.29                             & 35.76                             & 32.06                                & 38.76                                \\
                                                    & PPL                                                & 42.68                             & 50.15                             & 60.12                             & 50.98                                & 4.85                             & 43.83                             & 32.14                             & 13.63                             & 41.80                             & 44.28                             & 30.09                                & 37.05                                \\
                                                    & Similarity                                         & 46.98                             & 48.98                             & 62.13                             & 52.70                                & 9.71                             & 40.58                             & 39.89                             & 13.63                             & 42.92                             & 40.08                             & 31.14                                & 38.32                                \\
                                                    & LESS                                               & 44.67                             & 51.12                             & 61.88                             & 52.56                                & 10.68                            & 41.22                             & 40.15                             & 15.18                             & 48.74                             & 39.15                             & \underline{32.52}                                & \underline{39.20}                                \\
\TableLineColor \cellcolor{white}                   &  \textbf{DIQ (ours)}                                & 53.50                             & 54.15                             & 66.76                             & \textbf{58.14}                       & 12.62                            & 45.45                             & 42.21                             & 13.80                             & 44.28                             & 40.35                             & \textbf{33.12}                       & \textbf{41.46}                       \\ \cmidrule(l){2-14} 
                                                    & \textit{10\%}                                      &                                   &                                   &                                   &                                      &                                  &                                   &                                   &                                   &                                   &                                   &                                      &                                      \\
                                                    & Random                                             & 51.14                             & 39.04                             & 45.27                             & 45.15                                & 16.50                            & 45.13                             & 42.50                             & 16.49                             & 42.89                             & 40.93                             & 34.07                                & 37.77                                \\
                                                    & PPL                                                & 50.98                             & 39.98                             & 46.88                             & \underline{45.95}                                & 11.65                            & 45.80                             & 49.61                             & 16.80                             & 43.18                             & 40.00                             & 34.51                                & \underline{38.32}                                \\
                                                    & Similarity                                         & 51.22                             & 40.00                             & 40.00                             & 43.74                                & 10.68                            & 47.34                             & 38.12                             & 13.18                             & 43.92                             & 39.33                             & 32.10                                & 35.98                                \\
                                                    & LESS                                               & 49.98                             & 36.18                             & 41.68                             & 42.61                                & 9.71                             & 47.70                             & 48.20                             & 18.00                             & 44.00                             & 44.57                             & \underline{35.36}                                & 37.78                                \\
\TableLineColor \cellcolor{white}\multirow{-13}{*}{FineMed~\cite{finemedlm}} &  \textbf{DIQ (ours)}                                & 51.61                             & 40.40                             & 45.91                             & \textbf{45.97}                       & 17.48                            & 48.05                             & 43.83                             & 18.57                             & 44.87                             & 43.55                             & \textbf{36.06}                       & \textbf{39.36}                       \\ \bottomrule
\end{tabular}
\end{adjustbox}
\caption{Downstream task performance comparison of Llama3.1-8B-Instruct fine-tuned under 1\% and 10\% data retention ratios. Our DIQ-selected subset is compared with training on the full dataset, subsets from baseline methods, and other general models for reference. We repeat each setting five times and the average results are reported. Avg$_S$, Avg$_C$, and Avg$_A$ are the average accuracies for standard, challenging, and all tasks. \textbf{Bold} values highlight the best performance and \underline{underlined} values highlight the second best performance.
}
\label{tab:main_llama31}
\end{table*}

\paragraph{Datasets.}

For training, we leveraged a collection of meticulously constructed medical reasoning datasets. 
This includes five medium-scale datasets, each comprising no more than 32k examples: Huatuo \cite{huatuogpto1}, Huatuo-DS \cite{huatuogpto1}, FineMed \cite{finemedlm}, MedReason \cite{medreason}, and m1 \cite{m1}. 
Additionally, we incorporated one large-scale dataset: UltraMedical \cite{ultramedical} which contains 410k samples. 
For evaluation, we conducted a comprehensive assessment across nine benchmark tasks, categorized into standard and challenging tests. 
The standard test set comprised MedQA (MedQ) \cite{medqa}, MedMCQA (MedM) \cite{medmcqa}, and the medical subset of MMLU-Pro \cite{mmlupro}. 
For the more challenging evaluation, we utilized the biomedical portion of HLE \cite{hle}, MedBullets-option4 (MeB4) \cite{medbullets}, MedBullets-option5 (MeB5) \cite{medbullets}, MedXpertQA (MedX) \cite{medxpertqa}, MedGUIDE (MedG) \cite{medguide}, and MetaMedQA (MetM) \cite{metamedqa}, ensuring broad coverage across medical domains and reasoning complexities.
\vspace{-1em}

\paragraph{Models.}

For training our models, we selected a range of cutting-edge LLMs, including Qwen3-8B/14B/32B \cite{qwen3}, and Llama3.1-8B-Instruct \cite{llama31}.
For comprehensive comparison and reference, we also evaluated a diverse set of LLMs, categorized as follows:
1) General non-reasoning models: GPT-4.1, DeepSeek-V3-0324 \cite{deepseek}, and Gemini-2.5-Flash.
2) General reasoning models: DeepSeek-R1-0528, QwQ-32B \cite{qwq}, o4-mini-medium \cite{gpt4}, and Gemini-2.5-Pro \cite{gemini25}.
\vspace{-1em}
\paragraph{Baselines.}

We compare DIQ against four baselines: (i) Random, which samples instances uniformly at random; (ii) PPL \cite{meta_rater}, which computes full-text perplexity and selects the bottom-$k$ instances (lowest perplexity); (iii) Similarity \cite{sbert}, which computes cosine similarity between candidate embeddings and those from $\mathcal{D}_{\text{val}}$ and selects the top-$k$ subset; and (iv) LESS \cite{less}, which ranks candidates using TracIn influence scores \cite{tracin} and selects the top-$k$ subset.

\paragraph{Implementation Details.}

We fine-tuned all selected models using LoRA \cite{lora} with the following hyperparameters: the LoRA target modules include the query, key, and value projections; the LoRA rank was set to 8; the maximum context length was 8,192 tokens; the learning rate was $1\times10^{-4}$ with a cosine decay schedule; and training was conducted for 3 epochs. 
All training runs were performed on a Ubuntu 22.04 server equipped with 4x NVIDIA A800 GPUs.
We report the average accuracy for each task. 
To accommodate formatting variability in model outputs, correctness is determined via a two-step process: (1) we first check for an exact match of the correct option; (2) if that fails, we then check for the presence of the ground-truth answer text within the generated content. 
An answer is deemed incorrect only if both steps fail.
Full results of all experiments in this paper are provided in App.~\ref{app:full_all}.

\subsection{Main Results}

\paragraph{DIQ matches or exceeds full fine-tuning with 1-10\% data.}

As shown in Tab.~\ref{tab:main_llama31}, DIQ is the best-at-budget selector across two datasets (Huatuo and FineMed): at both 1\% and 10\% keep ratios it yields the highest Avg$_S$, Avg$_C$, and Avg$_A$ among all baseline methods.
Concretely, on Huatuo, DIQ improves Avg$_A$ from 40.54 (LESS, 1\%) and 41.73 (Similarity, 10\%) to 42.78 and 44.04 (+2.24 and +2.31). 
On FineMed, DIQ raises Avg$_A$ from 39.20 (LESS, 1\%) and 38.32 (PPL, 10\%) to 41.46 and 39.36 (+2.26 and +1.04).
Beyond beating the baselines, DIQ is highly data efficient: using only 10\% data, it \textit{surpasses full-data fine-tuning} on both datasets (Huatuo: 44.04 vs.\ 43.44; FineMed: 39.36 vs.\ 38.57), and with only 1\% data on FineMed it even outperforms the full-data model (41.46 vs. 38.57).
On Huatuo at 1\%, DIQ nearly matches the full-data result (42.78 vs. 43.44, -0.66) while using 99\% fewer samples.
The gains are especially pronounced on standard tasks: for FineMed, DIQ at 1\% lifts Avg$_S$ by +15.62 over full-data training (58.14 vs.\ 42.52), and for Huatuo at 10\% it improves Avg$_S$ over full data by +3.34 (58.11 vs.\ 54.77).
Improvements on challenging tasks are smaller but consistent at the same budget (e.g., +2.44 Avg$_C$ on Huatuo 1\% and +0.70 on FineMed 10\% over the best baselines), indicating that DIQ preserves hard-case coverage while pruning redundant samples.

\begin{figure*}[!t]
    \centering
    \includegraphics[width=0.96\linewidth]{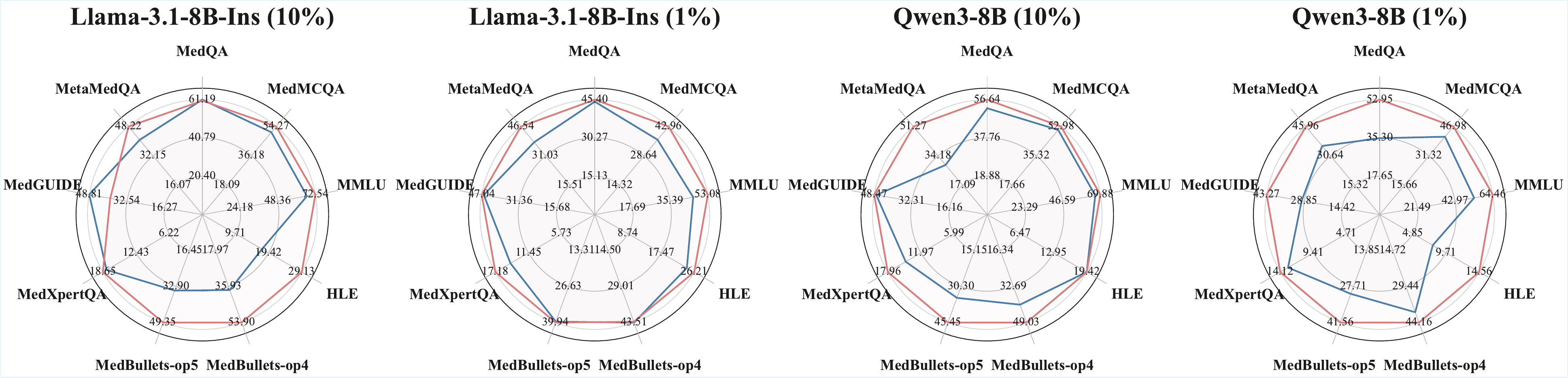}
    \caption{Downstream task performance comparison of models trained on MedReason-QA data selected from different methods at different data keeping ratios. Red line denotes DIQ, and blue line denotes random selection.}
    \label{fig:medreason_qa}
\end{figure*}

\paragraph{DIQ is also effective for QA datasets.}

To isolate the effect of DIQ when complex intermediate reasoning is absent, we construct MedReason-QA by removing CoT traces from MedReason \cite{medreason} while keeping the original splits and answer keys. 
As shown in Fig.~\ref{fig:medreason_qa}, DIQ consistently outperforms random selection at both 1\% and 10\% data budgets for Llama3.1-8B-Instruct and Qwen3-8B, with averages computed over five runs. 
This indicates that DIQ does not rely on explicit reasoning traces: even under QA-only supervision, where labels are short and signal-to-noise is lower, DIQ still identifies high-information density samples and preserves coverage of rare and clinically important cases. 
Practically, when CoT annotations are unavailable or costly, DIQ serves as a drop-in data selector for medical QA datasets, delivering robust gains under tight budgets. 
Full results are reported in App.~\ref{app:full_qa}.

\paragraph{DIQ enriches clinically salient reasoning signals.}

To examine whether DIQ favors samples that matter in clinical practice, three experienced clinicians distilled a rubric after reviewing 50 model rationales: \textit{Differential Diagnosis (DDx)}, \textit{Safety Check}, and \textit{Evidence Citation}.
We then ran an LLM-as-judge evaluation with Gemini-2.5-pro (5-point scale; details in App.~\ref{app:clinical_value}) along two dimensions: (i) \textit{data-level} quality by scoring $n=100$ Huatuo instances from the DIQ 1\% subset versus $n=100$  from the remainder; and (ii) \textit{model-level} clinical reasoning by scoring $n=100$ MedQA rationales from a Qwen3-8B model trained on DIQ-1\% versus $n=100$ from a model trained on the full Huatuo.
As shown in Tab.~\ref{tab:clinical_value}, DIQ-1\% is consistently higher on all three clinical metrics. 
At the data level, the improvements are +0.80, +0.35, +0.46, respectively; at the model level, gains persist with +0.05, +0.16 and +0.15. 
The strongest lift on \textit{DDx} suggests DIQ surfaces cases with richer hypothesis enumeration and risk triage, while positive shifts on Safety/Evidence indicate better red-flag screening and guideline grounding. 
These results connect DIQ’s quantitative gains to clinically meaningful behaviors, even when training on only 1\% of the data.
One case study is provided in App.~\ref{app:case_study}.

\begin{table}[!tb]
\centering
\small
\begin{tabular}{lccc}
\toprule
\textbf{Data}      & \textbf{DDx}  & \textbf{SC} & \textbf{EC} \\ \midrule
Full                & 3.59          & 3.33                  & 4.31                   \\
\TableLineColor 1\% DIQ             & \textbf{4.39} & \textbf{3.68}         & \textbf{4.77}  \\ \midrule
Full Generation     & 3.66          & 3.14                  & 4.75                   \\
\TableLineColor 1\% DIQ Generation & \textbf{3.71} & \textbf{3.30}         & \textbf{4.90}  \\ \bottomrule
\end{tabular}
\caption{Clinical value comparison: DIQ-1\% Data vs. remainder, and DIQ-1\% model vs. full-dataset model. All three evaluation metrics are rated on a 5-point scale. Abbreviations: DDx = Differential Diagnosis, SC = Safety Check, EC = Evidence Citation.}
\label{tab:clinical_value}
\end{table}
\section{Analysis}
\label{sec:analysis}

\begin{figure}[!t]
    \centering
    \includegraphics[width=0.8\linewidth]{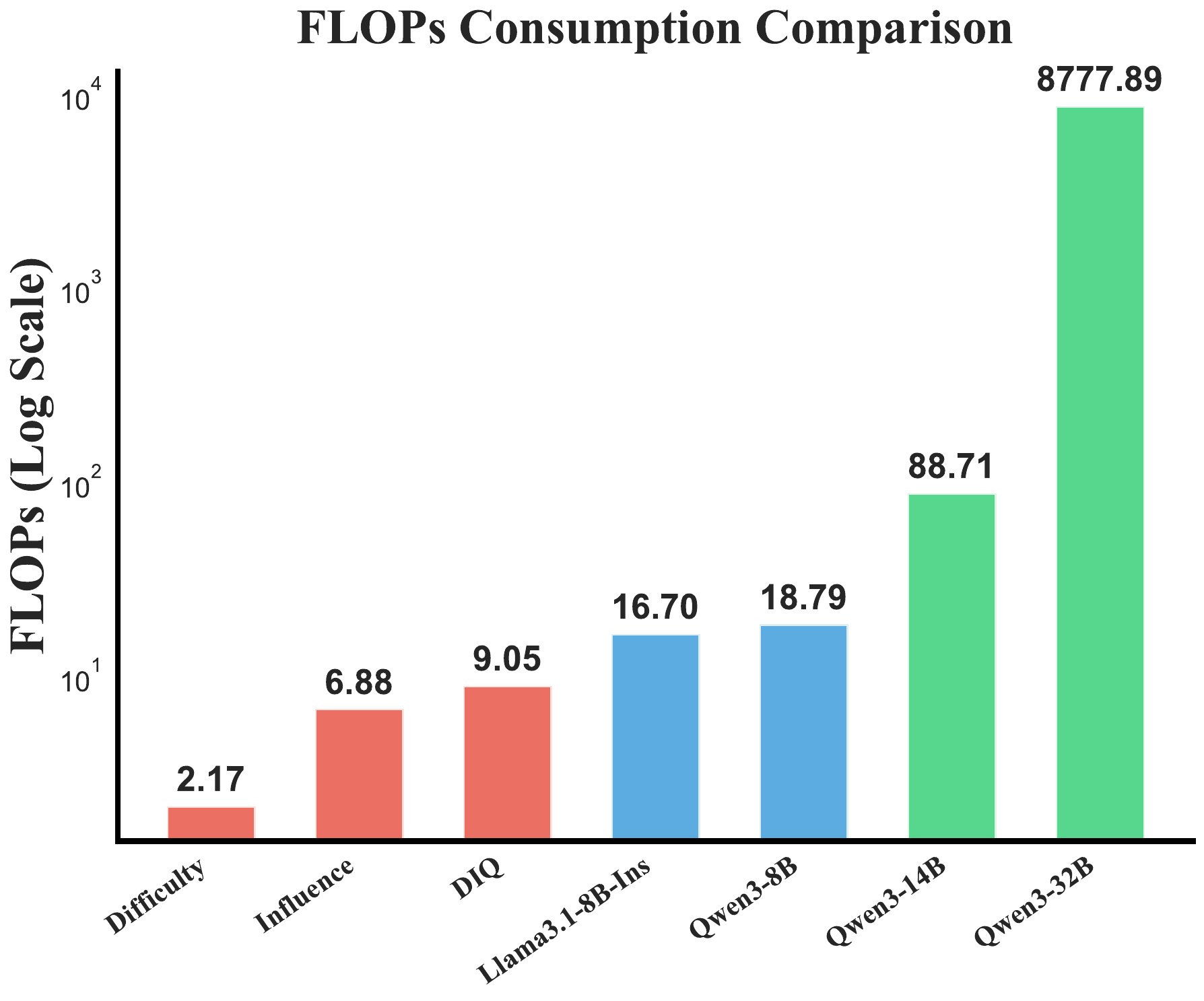}
    \caption{The FLOPs consumption ($10^{14}$) comparison of computing DIQ scores, fine-tuning Llama3.1 and Qwen3 series models. The y-axis is log scale for better presentation.}
    \label{fig:efficiency}
\end{figure}

\begin{figure*}[!t]
    \centering
    \includegraphics[width=1.0\linewidth]{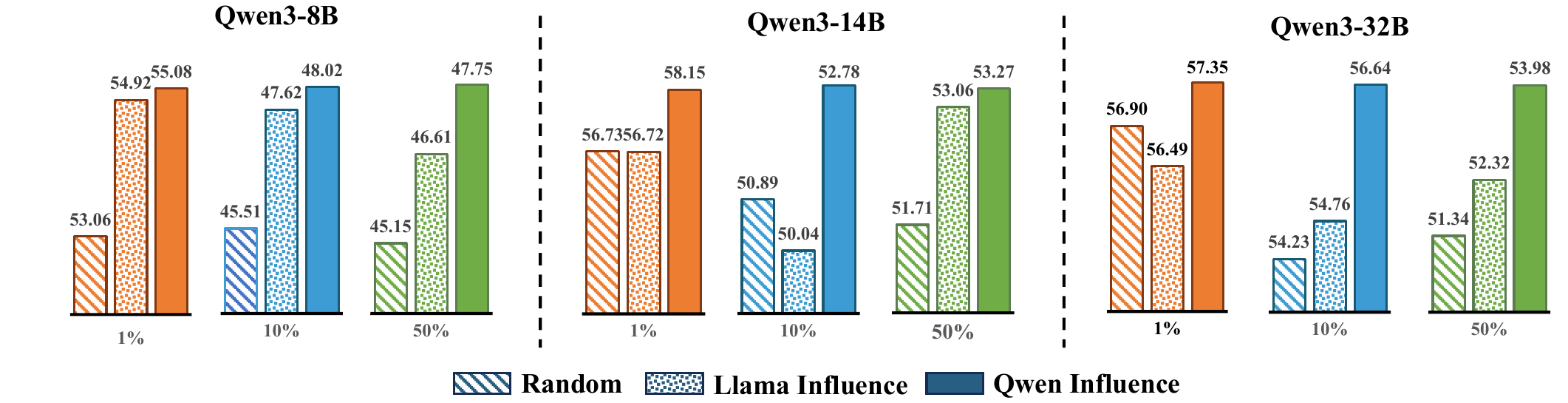}
    \caption{Downstream task performance of Qwen3-series models trained on DIQ-selected Huatuo  at different influence scores.}
    \label{fig:influence_analysis}
\end{figure*}

\begin{figure}[!t]
    \centering
    \includegraphics[width=\linewidth]{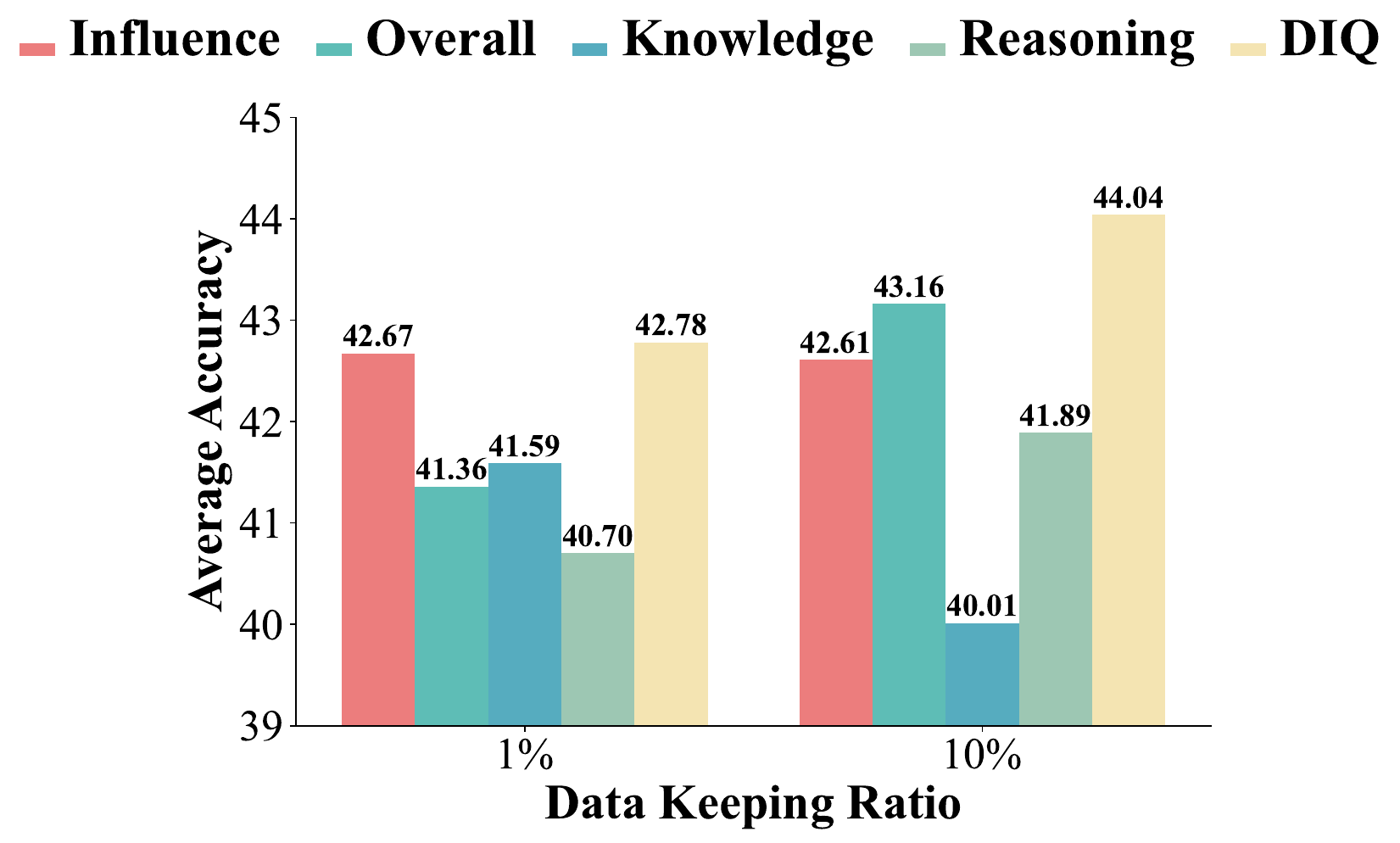}
    \caption{Average accuracy across all tasks of Llama3.1-8B-Instruct under different ablation settings.}
    \label{fig:ablation}
\end{figure}

\subsection{Efficiency Analysis}

A key advantage of DIQ is its \textit{one-time, model-agnostic} selection cost. 
As shown in Fig.~\ref{fig:efficiency}, running DIQ on Huatuo costs \textbf{9.05} (normalized FLOPs; details in App.~\ref{app:cost}), which is \textbf{1.85x} cheaper than a single full-data fine-tuning of Llama3.1-8B-Instruct (16.70) and \textbf{2.08x} cheaper than Qwen3-8B (18.79). 
The gap widens for larger models: DIQ is \textbf{9.80x} cheaper than a run costing 88.71 and nearly \textbf{970x} cheaper than one at 8777.89. 
In practice, this makes the selection pass a small fraction of end-to-end budget even before any reuse.
Moreover, the computed difficulty and influence scores can be cached and reused across multiple fine-tuning experiments, making the initial investment negligible, especially when training various and multiple models or tuning hyperparameters.
More details of FLOPs computation are provided in App.~\ref{app:cost}.

\subsection{Ablation Study}
\label{sec:ablation}

We ablate the contribution of each selection signal by comparing DIQ with single-criterion selectors that keep the top-$k$ examples ranked by either the influence score or one difficulty dimension (\textit{Knowledge}, \textit{Reasoning}, \textit{Overall}): under identical fine-tuning budgets, we train Llama3.1-8B-Instruct on the resulting subsets and report average accuracy at 1\% and 10\%. 
As shown in Fig.~\ref{fig:ablation}, DIQ attains the highest accuracy at both retention levels (42.78\% at $1\%$ and 44.04\% at $10\%$), surpassing the strongest single-criterion (Reasoning-only) by +0.89 and +0.88 points, respectively; averaged over all single-criterion baselines, the margins are +1.73 (1\%) and +1.59 (10\%). 
These results indicate that over-indexing on any single attribute, difficulty or influence, is brittle, while balancing complementary through DIQ signals yields more robust subsets under the same compute.

\subsection{Effect of Validation Set Size}

DIQ relies on a held-out validation set to estimate example influence and calibrate difficulty, so its size trades off estimator stability against compute. As shown in Table~\ref{tab:val_size_ablation} (Llama3.1-8B-Instruct on Huatuo at a 1\% keeping ratio), enlarging the validation split from 90 to 450 yields steady overall gains (Avg$_A$: 42.04 $\rightarrow$ 43.89, +1.85). Improvements are not linear: most of the benefit is realized by 360 examples (Avg$_A$=43.80, +1.76 over 90), after which returns are marginal (+0.09 at 450). By dimension, Avg$_S$ grows nearly monotonically (56.03 $\rightarrow$ 57.76), whereas Avg$_C$ is mildly non-monotonic—peaking at 360 (36.99) and remaining numerically comparable at 450 (36.95, $\Delta$=0.04). These trends suggest that moderate validation sizes already stabilize DIQ’s influence rankings while avoiding the extra compute of very large splits; in practice, 360--450 examples provide a strong quality-cost trade-off under this budget.

\begin{table}[tb]
\centering
\small
\begin{tabular}{cccc}
\toprule
\textbf{Validation Set Size} & \textbf{Avg$_S$} & \textbf{Avg$_C$} & \textbf{Avg$_A$} \\ \midrule
90                                               & 56.03            & 35.05            & 42.04            \\
180                                              & 56.54            & 35.91            & 42.78            \\
270                                              & 57.36            & 35.78            & 42.98            \\
360                                              & \underline{57.42}            & \textbf{36.99}   & \underline{43.80}            \\
450                                              & \textbf{57.76}   & \underline{36.95}            & \textbf{43.89}   \\ \bottomrule
\end{tabular}
\caption{Downstream task performance of Llama3.1-8B-Instruct models trained on Huatuo at 1\% keeping ratio under different validation set size settings of DIQ.}
\label{tab:val_size_ablation}
\end{table}

\subsection{Generalization of DIQ}

\paragraph{DIQ generalizes to cross-scale and cross-family models.}

To test whether DIQ’s influence-guided selection transfers beyond the model on which influence is computed, we compute influence once on a source model and reuse it to form DIQ subsets for larger targets within the same family and for models from a different family, with results shown in Fig.~\ref{fig:influence_analysis}. 
Within-family transfer (Qwen3-8B $\rightarrow$ Qwen3-14B/32B) is consistently strong: on Qwen3-14B, DIQ with Qwen-sourced influence improves over random by +1.42/+1.89/+1.56 points at 1\%/10\%/50\% retention, and on Qwen3-32B by +0.45/+2.41/+2.64, respectively. 
Cross-family transfer (Llama3.1-8B influence applied to Qwen3 targets) remains beneficial in 6/9 settings, with gains up to +2.11 (Qwen3-8B under 10\%), but can be neutral or slightly negative at the lowest budgets on larger targets (e.g., $-0.01$ on Qwen3-14B under 1\%, $-0.41$ on Qwen3-32B under 1\%, $-0.85$ on Qwen3-14B under 10\%). 
Taken together, these results indicate that DIQ’s influence component generalizes well across scale, while cross-family transfer is feasible but budget-sensitive, consistent with a rank-alignment view of influence and suggesting that mixing family-specific influence or reweighting by difficulty could further reduce gap.

\paragraph{DIQ for Preference Learning.}

We assess whether selective SFT via DIQ helps or harms downstream preference alignment by applying Direct Preference Optimization (DPO) \cite{dpo} on the FineMed preference corpus to every SFT model, and comparing against full-data baselines (Table~\ref{tab:rlhf}). 
DIQ-selected subsets are competitive or superior after DPO: relative to the full-data SFT\,+\,DPO baseline (Avg$_A$=55.52), {1\% DIQ + DPO achieves the best overall score (Avg$_A$ = 56.52; +1.00), 10\% DIQ + DPO also surpasses it (55.70; +0.18), and 50\% DIQ + DPO remains on par (55.37; $-0.15$). 
At matched retention budgets, DIQ consistently outperforms random selection on Avg$_A$ (+0.44/+0.73/+0.70 at 50\%/10\%/1\%) and on Avg$_C$ (+0.71/+1.19/+0.82), with neutral-to-small changes on Avg$_S$ ($-0.09$/$-0.19$/+0.45). 
These results indicate that curating SFT with DIQ produces a stronger model for preference learning: the gains not only survive the DPO stage but can exceed training on the full SFT corpus, highlighting an efficient and scalable path toward alignment in complex medical reasoning scenarios.

\begin{table}[!t]
\centering
\small
\begin{tabular}{lccc}
\toprule
\textbf{Setting} & \multicolumn{1}{l}{\textbf{Avg$_S$}} & \multicolumn{1}{l}{\textbf{Avg$_C$}} & \multicolumn{1}{l}{\textbf{Avg$_A$}} \\ \midrule
FineMed                                  & 73.29 & 45.92 & 55.04          \\ 
FineMed + DPO                            & 74.83 & 45.87 & 55.52          \\ \midrule 
50\% Random + DPO                        & 74.87 & 44.96 & 54.93          \\
\TableLineColor
50\% DIQ + DPO                           & 74.78 & \textbf{45.67} & \textbf{55.37} \\ \midrule 
10\% Random + DPO & 74.70 & 45.10 & 54.97          \\
\TableLineColor
10\% DIQ + DPO                           & 74.51 & \textbf{46.29} & \textbf{55.70} \\ \midrule
1\% Random + DPO  & 74.74 & 46.36 & 55.82          \\
\TableLineColor
1\% DIQ + DPO                            & \textbf{75.19} & \textbf{47.18} & \textbf{56.52} \\ \bottomrule
\end{tabular}
\caption{Downstream performance comparison of trained Qwen3-8B models using SFT and DPO.}
\label{tab:rlhf}
\end{table}
\section{Conclusion}
In this paper, we introduce DIQ, a data selection framework that identifies compact yet high-value training subsets for medical reasoning by jointly considering sample \textit{difficulty} and model-dependent \textit{influence} via a simple gradient inner product (\textit{Dot}). 
DIQ maps each example into the difficulty-influence plane and prioritizes the high-difficulty/influence region, yielding a principled, model-aware curriculum. 
Across nine downstream tasks, fine-tuning on only 1--10\% of DIQ-selected data matches or surpasses training on the full dataset, consistently preserving or improving accuracy while substantially reducing the training data footprint and computational cost. 

Due to computational resource constraints, we have not yet evaluated DIQ on very large LLMs (e.g., $\ge$70B parameters). 
In future work, we will scale up our study to such models to characterize how DIQ’s gains evolve with model capacity. 
We will also explore dynamic variants that periodically refresh influence estimates during training and assess robustness to the choice of validation sets and to broader clinical subdomains and application scenarios.

\section*{Acknowledgement}

This work is supported by the National Key Research and Development Program of China under Grant 2024YFC3308500.

{
    \small
    \bibliographystyle{ieeenat_fullname}
    \bibliography{main}
}

\appendix
\clearpage
\setcounter{page}{1}
\maketitlesupplementary

\startcontents[appendixtoc]
\printcontents[appendixtoc]{l}{1}{
    \setcounter{tocdepth}{2}
    \section*{Appendix Contents}
}

\section{Details of Pilot Experiment}
\label{app:pilot}

To investigate the interplay between medical difficulty and sample influence, we conducted a pilot experiment on the FineMed dataset. 
We first partitioned the data into four quadrants (We set difficulty score threshold as 3 and choose the median score of influence scores as bounder) based on difficulty and influence scores: $\mathcal{Q}_1$ with high difficulty and high influence, $\mathcal{Q}_2$ with low difficulty and high influence, $\mathcal{Q}_3$ with high difficulty and low influence, and $\mathcal{Q}_4$ with low difficulty and low influence. 
Subsequently, we fine-tuned separate instances of the Qwen3-8B model, each using a 1\% data subset drawn exclusively from one of the four quadrants. 
The impact of each quadrant was assessed through a two-pronged evaluation: 1) qualitative reasoning ability, scored on a 5-level Likert scale by Gemini-2.5-pro, and 2) quantitative task performance, measured by accuracy across nine downstream datasets. 
The full results of pilot experiment are shown in Table \ref{tab:full_pilot}. 
The prompt for the reasoning quality evaluation is provided below.

\begin{tcolorbox}[promptboxstyle, title=Reasoning Quality]
You are an experienced medical doctor and your task is to systematically evaluate and score the clinical reasoning process.

\textbf{I. Aspects to Consider for Evaluation}

When reading and analyzing a medical reasoning text, please consider the following three core areas holistically:

1. Analysis and Reasoning Process

Completeness of Information: Was all key clinical information (history, signs, lab and imaging results, etc.) accurately and comprehensively identified?

Synthesis of Information: Was scattered data (symptoms, risk factors, test results) effectively synthesized into a coherent and meaningful clinical picture?

Logical Chain: Is the reasoning process clear, rigorous, and progressive? Are there any logical leaps or contradictions?

Differential Diagnosis: Were other relevant possibilities (key differential diagnoses) considered and reasonably ruled out based on the available evidence?

2. Application of Knowledge

Accuracy of Knowledge: Is the applied medical knowledge (e.g., pathophysiology, epidemiology, drug mechanisms) accurate?

Adherence to Guidelines: Does the understanding of diagnostic criteria and treatment options align with current, accepted clinical guidelines and evidence-based medicine?

3. Conclusion and Justification

Correctness of Conclusion: Is the final diagnosis and proposed management plan correct?

Quality of Justification: Is the reasoning provided for the final conclusion clear, persuasive, and well-supported by the evidence in the case?

\textbf{II. Comprehensive Scoring Rubric (1-5 Points)}

After holistically considering all the points above, assign a single comprehensive score that best reflects the overall quality, based on the following criteria:

5 (Excellent):
The reasoning process is exemplary. The analysis is thorough, the logic is flawless, the application of knowledge is precise, and the conclusion is correct and exceptionally well-justified. It mirrors the thinking of an expert clinician.

4 (Good):
The reasoning process is strong and leads to the correct conclusion. The core logic and knowledge are sound, but there may be minor omissions in how the process is presented (e.g., not fully elaborating on the differential diagnosis), without affecting the overall outcome.

3 (Adequate):
The reasoning arrives at the correct conclusion, but the process has noticeable shortcomings. The logical chain may be unclear, the justification weak, or it may rely more on "pattern matching" than systematic analysis. It answers "what" but not "how" or "why."

2 (Poor):
The reasoning process has significant flaws. It may miss key data, apply incorrect knowledge, or follow a convoluted logical path, often leading to an incorrect or incomplete conclusion.

1 (Very Poor):
The reasoning is fundamentally flawed, demonstrating a lack of basic understanding of the clinical scenario, significant knowledge errors, and a complete absence of logical structure. The conclusion is unsubstantiated.

Please use the following format for your response.

\begin{itemize}
    \item  Score: [1-5]
    \item  Rationale: [Provide a brief, specific justification for the score, citing examples from the response.]
\end{itemize}

Here are the Question and Answer: 

\end{tcolorbox}

\begin{table*}[!t]
\centering
\small
\begin{tabular}{@{}lccccccccccccc@{}}
\toprule
\textbf{Model} &
  \multicolumn{1}{l}{\textbf{MedQ}} &
  \multicolumn{1}{l}{\textbf{MedM}} &
  \multicolumn{1}{l}{\textbf{MMLU}} &
  \multicolumn{1}{l}{\textbf{Avg$_S$}} &
  \multicolumn{1}{l}{\textbf{HLE}} &
  \multicolumn{1}{l}{\textbf{MeB4}} &
  \multicolumn{1}{l}{\textbf{MeB5}} &
  \multicolumn{1}{l}{\textbf{MedX}} &
  \multicolumn{1}{l}{\textbf{MedG}} &
  \multicolumn{1}{l}{\textbf{MetM}} &
  \multicolumn{1}{l}{\textbf{Avg$_C$}} &
  \multicolumn{1}{l}{\textbf{Avg$_A$}} &
  \textbf{\begin{tabular}[c]{@{}c@{}}Reasoning \\ Quality\end{tabular}} \\ \midrule
1\% $\mathcal{Q}_1$ & 78.01 & 64.89 & 83.93 & \textbf{75.61} & 15.53 & 66.56 & 59.09 & 18.69 & 55.07 & 62.78 & \textbf{46.29} & \textbf{56.06} & \textbf{4.82} \\
1\% $\mathcal{Q}_2$ & 76.04 & 56.30 & 83.93 & \underline{72.09}          & 9.71  & 64.94 & 58.12 & 17.88 & 54.59 & 59.72 & \underline{44.16}          & \underline{53.47}          & 4.27          \\
1\% $\mathcal{Q}_3$ & 66.85 & 65.26 & 83.10 & 71.74          & 9.71  & 62.34 & 57.79 & 14.53 & 54.27 & 58.99 & 42.94          & 52.54          & \underline{4.60}          \\
1\% $\mathcal{Q}_4$ & 60.53 & 53.79 & 74.10 & 62.81          & 12.62 & 62.23 & 58.44 & 12.78 & 48.24 & 58.49 & 42.13          & 49.02          & 4.18          \\ \bottomrule
\end{tabular}
\caption{Full downstream task accuracy and reasoning quality results of pilot experiment.}
\label{tab:full_pilot}
\end{table*}

\section{Case Study}
\label{app:case_study}

As shown in Figure \ref{fig:case_study}, we provide a case study of Qwen3-8B trained on DIQ-1\% FineMed answering a question in MedBullets-option5 and mark the parts of \textit{Differential Diagnosis (DDx)}, \textit{Safety Check}, and \textit{Evidence Citation} in \textcolor{red}{Red}, \textcolor{orange}{Orange}, and \textcolor{blue}{Blue}.
\par
The model employs a systematic and evidence-based approach to clinical problem-solving. 
It initiates its analysis by correlating the patient's history and risk factors with key laboratory findings, principally the profoundly low CD4$^+$ count, to establish a diagnosis of severe immunosuppression. This correctly frames the presenting problem within the context of an opportunistic central nervous system infection.
Subsequently, the model focuses on the most diagnostically salient evidence from the lumbar puncture. 
It interprets the cerebrospinal fluid (CSF) profile—characterized by lymphocytic pleocytosis, hypoglycorrhachia (low glucose), and elevated protein—as highly suggestive of a fungal etiology. The positive India ink stain is correctly identified as the definitive finding that confirms a diagnosis of cryptococcal meningitis.
Finally, in determining the management plan, the model assesses the disease's severity. It logically selects the standard-of-care induction therapy for severe cryptococcosis, Amphotericin B and flucytosine, while correctly distinguishing this from treatments for other pathogens or from therapies, such as fluconazole, which are reserved for less severe presentations or consolidation phases.

\section{Details of Difficulty Score}
\label{app:difficulty_classifier}

\subsection{Prompt for Difficulty Score Annotation}

We provide the prompt for obtaining medical difficulty scores along three dimensions (\textit{Knowledge}, \textit{Reasoning}, and \textit{Overall}) in the following box.

\begin{tcolorbox}[promptboxstyle, title=Medical Difficulty]
You are an experienced medical doctor and independent practitioner. Your task is to classify a medical question across THREE dimensions following a specific evaluation sequence: First assess Knowledge Complexity, then Reasoning Complexity, and finally provide an Overall Difficulty rating that synthesizes both dimensions.

Evaluation Sequence: Knowledge → Reasoning → Overall

Please evaluate each dimension independently in the specified order, as this sequence ensures a more systematic and comprehensive assessment.

---

\subsection*{Dimension 1: Knowledge Complexity (1-5 Levels)}

Classify based on the depth and breadth of medical knowledge required:

\textbf{Level 1 (Basic Medical Knowledge):} The question requires fundamental medical concepts taught in early medical education. Common diseases, basic anatomy/physiology, standard definitions.

\textbf{Level 2 (Standard Clinical Knowledge):} The question requires typical clinical knowledge expected of practicing physicians. Common clinical presentations, standard diagnostic criteria, routine management principles.

\textbf{Level 3 (Specialty Foundational Knowledge):} The question requires specialized knowledge within specific medical fields. Subspecialty concepts, advanced pathophysiology, specialized diagnostic approaches.

\textbf{Level 4 (Deep Specialty Knowledge):} The question requires expert-level knowledge within specialized domains. Rare diseases, complex pathophysiology, advanced subspecialty management, cutting-edge diagnostic techniques.

\textbf{Level 5 (Cutting-edge/Rare Specialized Knowledge):} The question requires knowledge of very rare conditions, latest research findings, experimental treatments, or highly specialized expert-level concepts that even specialists might need to reference.

---

\subsection*{Dimension 2: Reasoning Complexity (1-5 Levels)}

Classify based on the level of medical reasoning difficulty required:

\textbf{Level 1 (Direct Recall/Understanding):} The question primarily tests direct recall of medical facts, definitions, common associations, or basic recognition. It requires no complex reasoning; the answer is a straightforward retrieval of memorized knowledge.

\textbf{Level 2 (Simple Application):} The question requires basic application of well-established medical knowledge to straightforward scenarios. Involves simple pattern recognition or direct application of standard protocols with minimal reasoning steps.

\textbf{Level 3 (Moderate Reasoning):} The question requires applying medical knowledge to specific, often slightly novel, scenarios. It involves interpreting clinical data, making logical connections between symptoms and conditions, or performing straightforward differential diagnosis. It typically involves 2-3 clear reasoning steps.

\textbf{Level 4 (Complex Reasoning):} The question demands integration of multiple pieces of information from various domains (e.g., history, physical, labs, imaging), complex differential diagnosis, evaluation of multiple management options, or navigating moderately ambiguous data. It involves multi-step logical chains and synthesis of information.

\textbf{Level 5 (Expert-level Reasoning/Complex Problem Solving):} The question requires advanced clinical reasoning with high-level integration of complex, ambiguous, or incomplete data from multiple domains. It involves sophisticated differential diagnosis, evaluation of competing hypotheses, critical evaluation of conflicting information, and navigation of highly nuanced clinical scenarios. Requires expert-level clinical judgment and complex multi-step reasoning chains.

\textbf{When determining reasoning level, consider:}
\begin{itemize}
    \item The amount of information provided in the question (how many data points need integration)
    \item The number and complexity of reasoning steps required
    \item The degree of ambiguity or nuance present in the scenario
    \item Whether the answer derives from direct recall versus requiring deductive/inductive reasoning
    \item The sophistication of clinical judgment required
\end{itemize}

---

\subsection*{Dimension 3: Overall Difficulty (1-5 Levels)}

Comprehensive assessment that synthesizes both Knowledge and Reasoning complexity:

\textbf{Level 1 (Very Easy):} Low knowledge requirements with minimal reasoning demands. Straightforward questions with clear answers, minimal clinical complexity, common scenarios.

\textbf{Level 2 (Easy):} Moderate knowledge requirements or simple reasoning, but not both simultaneously. Slightly more complex but still manageable scenarios.

\textbf{Level 3 (Moderate):} Balanced combination of knowledge and reasoning demands, or high complexity in one dimension compensated by lower complexity in the other. Moderate clinical complexity requiring integrated thinking.

\textbf{Level 4 (Hard):} High demands in both knowledge and reasoning, or extreme complexity in one dimension. Complex scenarios requiring advanced clinical judgment, significant ambiguity, multiple competing factors.

\textbf{Level 5 (Very Hard):} Exceptional demands in both knowledge and reasoning simultaneously. Extremely challenging scenarios requiring expert-level judgment, high ambiguity, multiple complex factors, potentially controversial or cutting-edge topics.

\textbf{Overall Difficulty Synthesis Guidelines:}
\begin{itemize}
    \item Consider how Knowledge and Reasoning complexity interact
    \item High knowledge + high reasoning = very challenging
    \item High knowledge + low reasoning = moderate challenge
    \item Low knowledge + high reasoning = moderate challenge
    \item Account for cumulative cognitive load
\end{itemize}

---

\subsection*{Output Format:}

Please provide your assessment in the following format:

Knowledge Complexity Score: [1-5]

Reasoning Complexity Score: [1-5]

Overall Difficulty Score: [1-5]

\textbf{Knowledge Justification:}
[Explain the knowledge requirements - medical domain depth, specialization level, rarity of concepts, specific medical knowledge needed]

\textbf{Reasoning Justification:}
[Explain the reasoning demands - information integration, logical steps, ambiguity handling, clinical reasoning complexity]

\textbf{Overall Difficulty Justification:}
[Explain how Knowledge and Reasoning complexity combine to create the overall challenge level, considering their interaction and cumulative impact]

\textbf{Key Factors:}
\begin{itemize}
    \item Primary difficulty drivers
    \item Interaction between knowledge and reasoning demands
    \item Clinical context considerations
    \item Any notable complexities or special considerations
\end{itemize}

Please evaluate the following medical reasoning question and note that you only need to evaluate the difficulty and you don't need to answer the question.
\end{tcolorbox}

\begin{figure*}
    \centering
    \begin{tcolorbox}[exampleboxstyle, title=Example Generated by Qwen3-8B trained on DIQ-1\% FineMed]
\textbf{Question:}

A 23-year-old woman presents to the emergency department complaining of a worsening headache. The headache started 1 month ago. It is constant and all over but gets worse when she is lying down or in the setting of bright lights. A review of systems is significant for a low-grade fever, night sweats, cough, malaise, poor appetite, and unintentional weight loss of 12 pounds in the last 2 months. The patient is sexually active with multiple male partners and reports inconsistent condom use. She has a history of intravenous drug use and has not been to a doctor in the last 2 years. Her temperature is $100.4\,^{\circ}\text{F}$ ($38.0\,^{\circ}\text{C}$), blood pressure is $110/78$ mmHg, pulse is $88/\text{min}$, and respirations are $14/\text{min}$ with an oxygen saturation of $98\,\%$ O$_2$ on room air. On physical exam, pain is elicited upon passive flexion of the patient’s neck. A CT scan shows ventricular enlargement. A CD4$^+$ count is $57~\text{cells}/\mu\text{L}$ blood. A lumbar puncture is performed with the following findings: \textbf{Opening pressure:} $210~\text{mmH}_2\text{O}$; \textbf{Glucose:} $32~\text{mg/dL}$; \textbf{Protein:} $204~\text{mg/dL}$;  \textbf{India ink stain:} Positive; \textbf{Leukocyte count and differential:} Leukocyte count: $200/\text{mm}^3$, Lymphocytes: $100\,\%$.
Which of the following treatments most directly addresses the underlying cause of this patient's symptoms?
\begin{enumerate}
    \item[A.] Acyclovir
    \item[B.] Amphotericin B and 5-flucytosine
    \item[C.] Dexamethasone
    \item[D.] Fluconazole
    \item[E.] Vancomycin and ceftriaxone
\end{enumerate}
\textbf{Answer:}
Okay, let's try to figure out this case. So, the patient is a 23-year-old woman with a worsening headache that's been going on for a month. The headache is constant, worse when lying down or in bright lights. She has a low-grade fever, night sweats, cough, malaise, poor appetite, and weight loss. She's sexually active with multiple partners and inconsistent condom use, and has a history of IV drug use. She hasn't seen a doctor in two years. Her vitals are temperature 100.4°F, BP 110/78, pulse 88, respirations 14, oxygen saturation 98\%. On exam, neck pain with passive flexion. CT shows ventricular enlargement. \textcolor{blue}{\textit{Evidence Citation:} CD4 count is 57 cells/µL. Lumbar puncture results: opening pressure 210 mmH2O, glucose 32 mg/dL, protein 204 mg/dL. India ink stain is positive. CSF leukocytes 200/mm³, all lymphocytes, RBCs 2/mm³.}
First, the key symptoms and findings. The headache, neck stiffness (Brudzinski sign?), fever, weight loss, and the \textcolor{blue}{\textit{Evidence Citation:} CD4 count is very low (57), which is indicative of advanced HIV.} The \textcolor{blue}{\textit{Evidence Citation:} India ink stain positive is a clue.} India ink is used to detect Cryptococcus neoformans, which is a fungal infection. \textcolor{red}{\textit{Differential Diagnosis (DDx):} But wait, in someone with a low CD4 count, they're immunocompromised, so they could have opportunistic infections. But Cryptococcus is a common cause of meningitis in immunocompromised patients, especially those with HIV.}
But the \textcolor{blue}{\textit{Evidence Citation:} CSF findings: elevated protein, low glucose, and India ink positive. So that's consistent with cryptococcal meningitis. The CSF leukocytes are 200, all lymphocytes, which is typical for cryptococcal meningitis.} The red blood cells in CSF might be due to traumatic lumbar puncture, but the number is low here. The opening pressure is high, which is common in meningitis.
\textcolor{orange}{\textit{Safety Check:} But wait, let me check the options again. So the CSF findings are consistent with cryptococcal meningitis. Therefore, the correct treatment would be B. Amphotericin B and 5-flucytosine. However, in some cases, fluconazole might be used as initial treatment, but in more severe cases, the combination is preferred. Also, the patient's CD4 count is very low, so she might need antiretroviral therapy as well, but the question is about the treatment for the underlying cause of the symptoms, which is the cryptococcal infection.} So B is the correct answer.
Wait, but let me think again. The \textcolor{blue}{\textit{Evidence Citation:} India ink stain is positive for Cryptococcus.} So yes, the treatment is amphotericin B and 5-flucytosine. So the answer is B.

    \end{tcolorbox}
    \caption{Case study of Qwen3-8B trained on DIQ-1\% FineMed answering a question in MedBullets-option5.}
    \label{fig:case_study}
\end{figure*}

\subsection{Difficulty Classifier Training}

We evaluated three lightweight BERT-style models for predicting medical difficulty: BiomedBERT \cite{biomedbert}, ClinicalBERT \cite{clinicalbert}, and ModernBERT \cite{modernbert}. 
As shown in Table \ref{tab:full_difficulty_classifier}, BiomedBERT consistently outperformed the other models and was therefore selected as our difficulty classifier.

\begin{table}[b]
\centering
\small
\begin{tabular}{@{}lccc@{}}
\toprule
\textbf{Difficulty} & \textbf{BiomedBERT} & \textbf{ClinicalBERT} & \textbf{ModernBERT} \\ \midrule
Knowledge            & \textbf{80.89} & 76.91 & 78.69 \\
Reasoning  & \textbf{83.86} & 82.69 & 83.55 \\
Overall              & \textbf{81.90} & 80.84 & 81.05 \\ \bottomrule
\end{tabular}
\caption{Test set F1 scores on difficulty classification task of three BERT-style models.}
\label{tab:full_difficulty_classifier}
\end{table}

\subsection{Difficulty Score Distribution}

We list the difficulty score distributions of FineMed, Huatuo, Huatuo-DS, and UltraMedical in Figures \ref{fig:diff_finemed}, \ref{fig:diff_huatuo}, \ref{fig:diff_huatuods}, \ref{fig:diff_m1}, and \ref{fig:diff_ultramedical}.

\begin{figure*}
    \centering
    \includegraphics[width=0.9\linewidth]{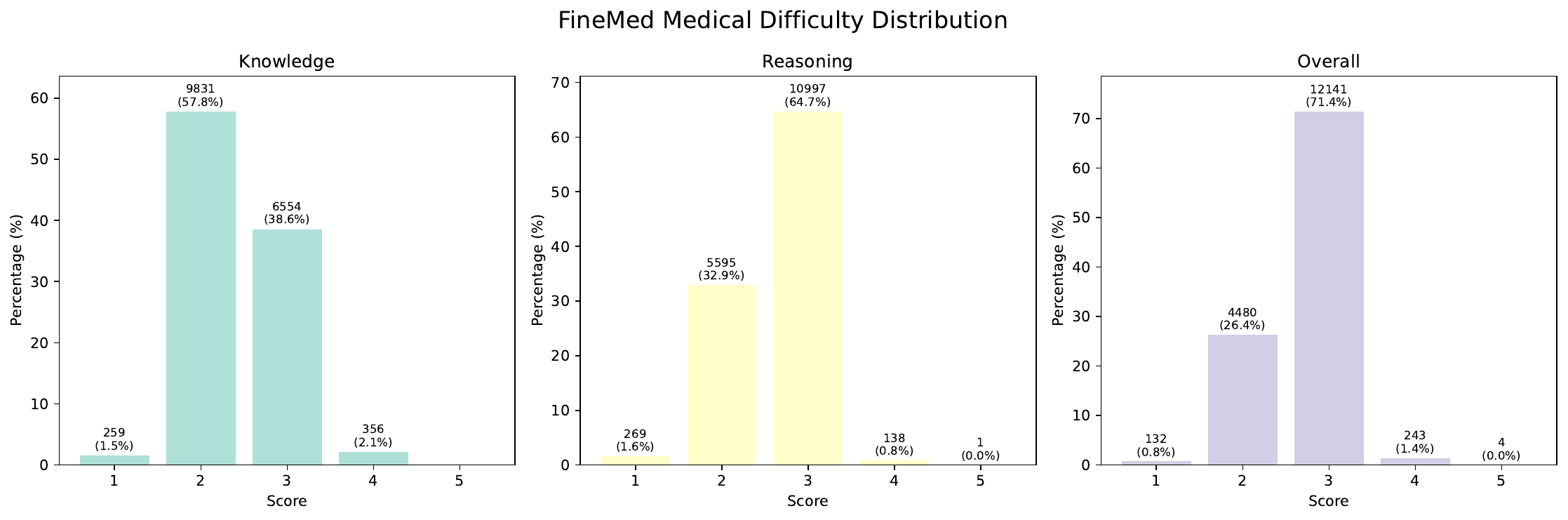}
    \caption{Difficulty score distribution of FineMed.}
    \label{fig:diff_finemed}
\end{figure*}

\begin{figure*}
    \centering
    \includegraphics[width=0.9\linewidth]{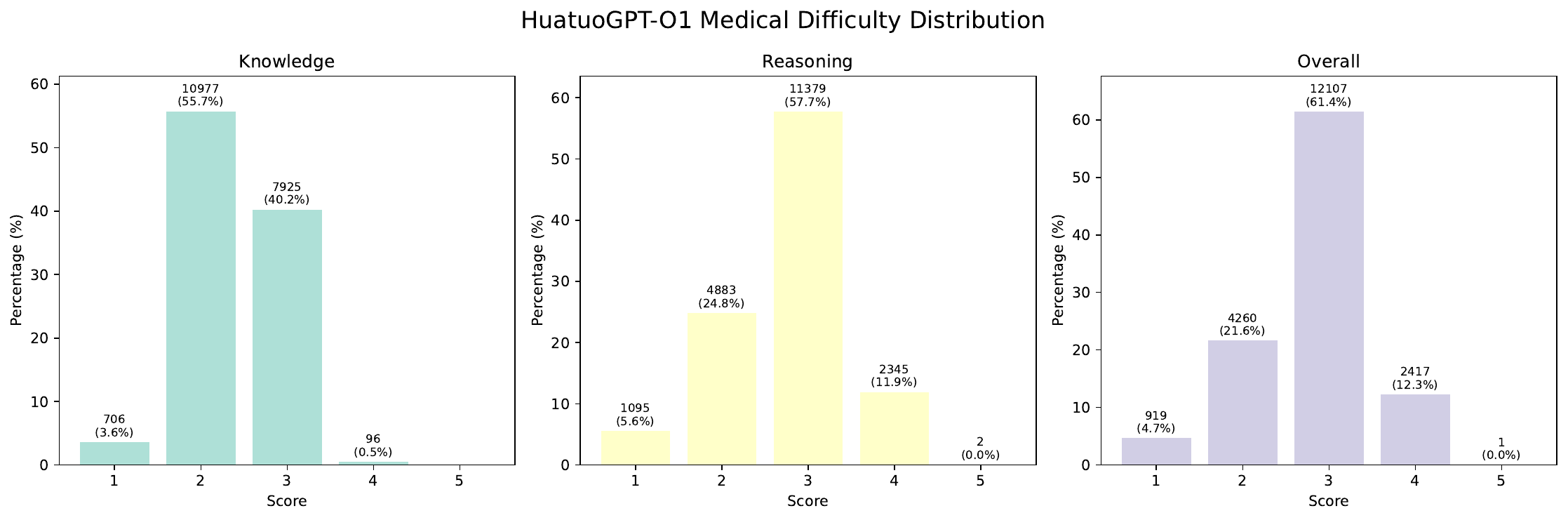}
    \caption{Difficulty score distribution of Huatuo.}
    \label{fig:diff_huatuo}
\end{figure*}

\begin{figure*}
    \centering
    \includegraphics[width=0.9\linewidth]{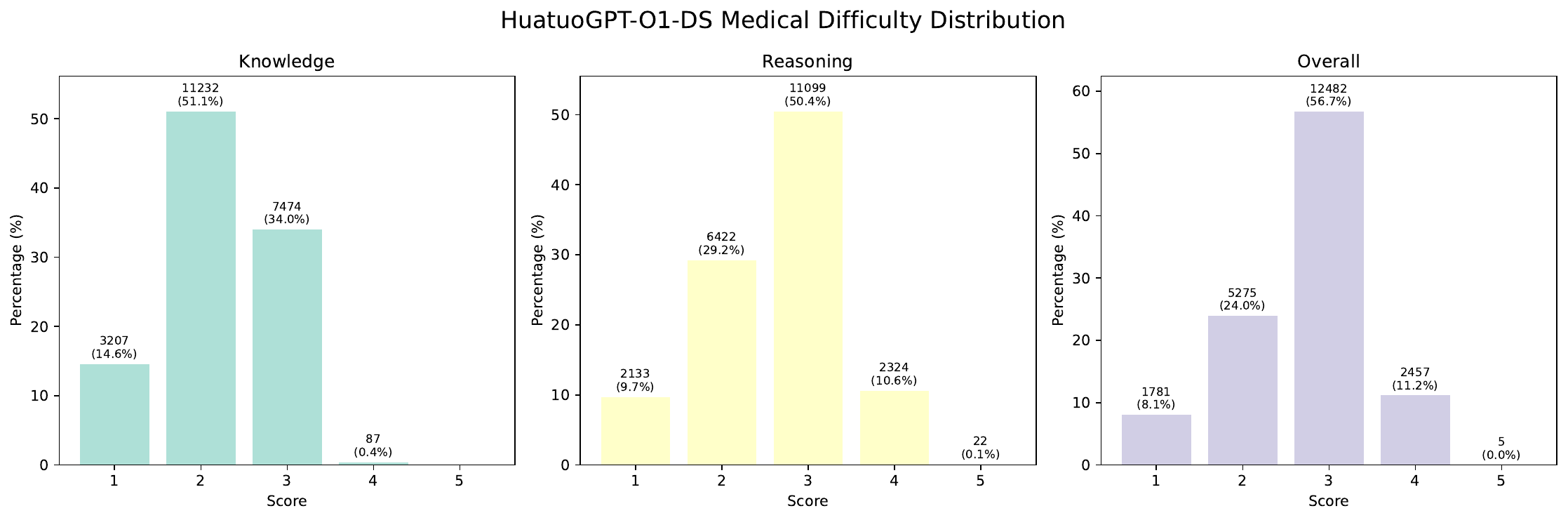}
    \caption{Difficulty score distribution of Huatuo-DS.}
    \label{fig:diff_huatuods}
\end{figure*}

\begin{figure*}
    \centering
    \includegraphics[width=0.9\linewidth]{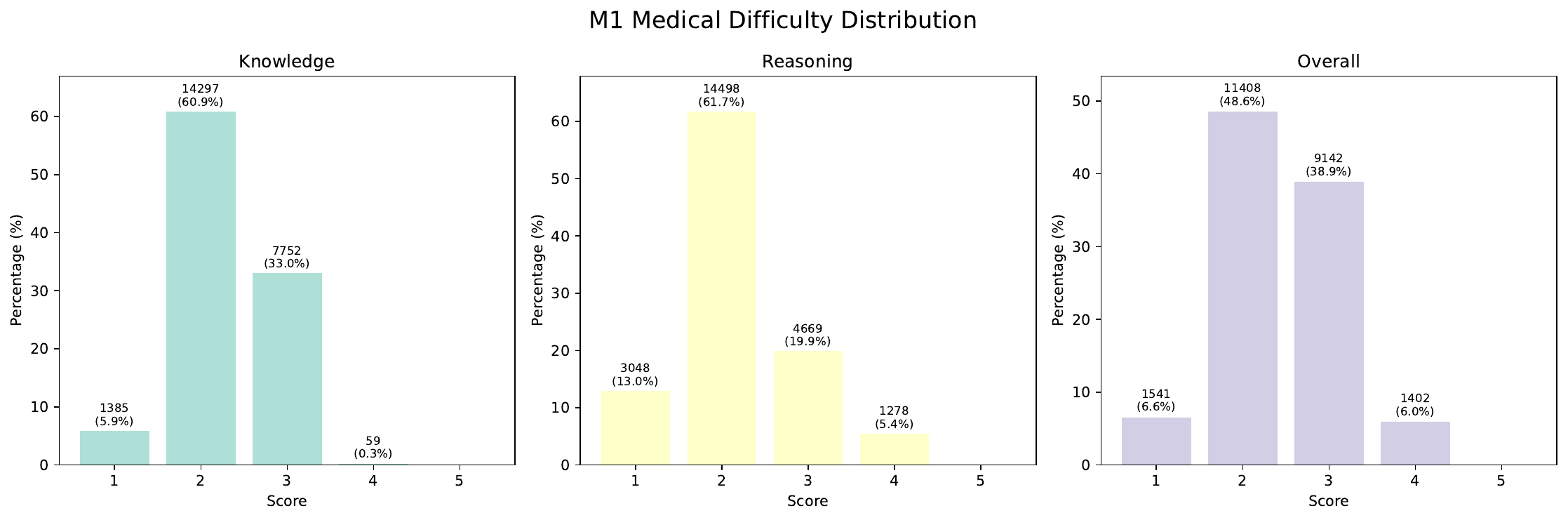}
    \caption{Difficulty score distribution of m1.}
    \label{fig:diff_m1}
\end{figure*}

\begin{figure*}
     \centering
    \includegraphics[width=0.9\linewidth]{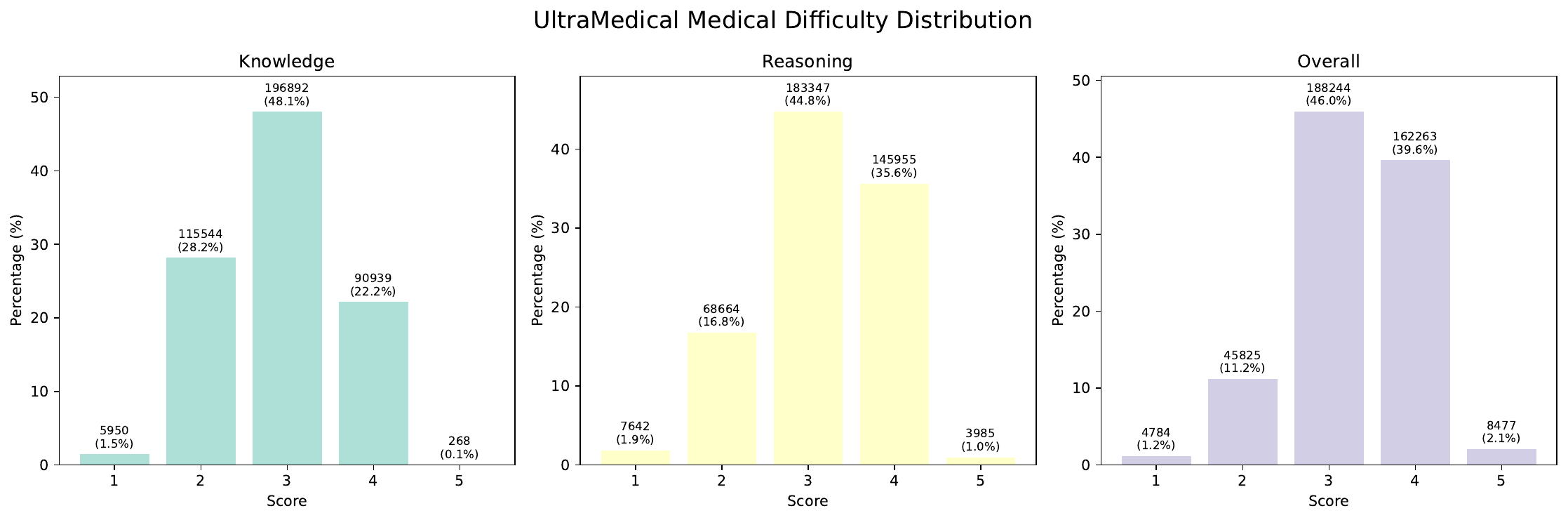}
    \caption{Difficulty score distribution of UltraMedical.}
    \label{fig:diff_ultramedical}
\end{figure*}

\section{Details of Influence Score}
\label{app:influence_score}

\subsection{Influence Score Computation}

We directly computed influence scores for medium-scale datasets (FineMed, Huatuo, Huatuo-DS, m1, and MedReason) using Equation \ref{eq:dot}. However, to conserve computational resources, for the large-scale UltraMedical dataset, we instead trained an influence rater to predict these scores, following a similar strategy to our difficulty classifier. 
For this regression task, we evaluated BiomedBERT, ClinicalBERT, and ModernBERT. 
The models were trained on a set of 45,000 instances and tested on 5,000 instances, all sampled from the medium-scale datasets. 
As shown in Table \ref{tab:full_influence_rater}, ModernBERT achieved the strongest performance and was selected as the influence rater.

\begin{table}[tb]
\centering
\small
\begin{tabular}{@{}lccc@{}}
\toprule
\textbf{Influence}   & \textbf{BiomedBERT} & \textbf{ClinicalBERT} & \textbf{ModernBERT} \\ \midrule
Llama3.1-8B-Ins & 73.92               & 76.92                 & \textbf{82.39}      \\
Qwen3-8B             & 79.92               & 80.06                 & \textbf{84.29}      \\ \bottomrule
\end{tabular}
\caption{Test set Spearman R scores on influence regression task of three BERT-style models.}
\label{tab:full_influence_rater}
\end{table}

\subsection{Influence Score Distribution}

We list the influence score distributions of FineMed, Huatuo, Huatuo-DS, m1, and MedReason in Figures \ref{fig:inf_finemed}, \ref{fig:inf_huatuo}, \ref{fig:inf_huatuods}, \ref{fig:inf_m1}, and \ref{fig:inf_medreason}.

\section{Details of Clinical Value Assessment}
\label{app:clinical_value}
To ground our evaluation in clinical practice, we consulted three experienced clinicians to review the reasoning processes generated by our models. 
This expert review identified three components as crucial for establishing clinical value: \textit{Differential Diagnosis (DDx)}, \textit{Safety Check}, and \textit{Evidence Citation}.
These components subsequently formed the basis for our automated evaluation prompt, which is provided below.

\begin{tcolorbox}[promptboxstyle, title=Clinical Value]
You are an experienced medical doctor and your task is to systematically evaluate and score the clinical reasoning process.  The evaluation is structured around three core clinical cognitive behaviors: Differential Diagnosis (DDx), Safety Check, and Evidence Citation.

\textbf{Instructions}:

- Read the medical Question and the Full Response: Carefully review the clinical scenario presented and the entire reasoning process.

- Evaluate Each Category Separately: For each of the three categories below, assess the performance against the described criteria.

- Assign a Score from 1 to 5: Use the detailed rubric to assign a score from 1 (Very Poor) to 5 (Excellent) for each category. Half-points (e.g., 3.5) are not permitted.

- Provide a Rationale: For each score, you must provide a brief, specific rationale explaining your decision. Justify the score by citing specific examples or omissions from the response.

- Use the Provided Output Template: Format your final evaluation using the template at the end of this document.

\textbf{Scoring Rubric}:

\textbf{1. Differential Diagnosis (DDx)}

This category assesses the ability to generate a list of potential diagnoses and systematically narrow it down using logical reasoning.

Level 5 (Excellent): The answer generates a comprehensive and relevant list of differential diagnoses, including both common and less common but critical possibilities. It systematically compares and contrasts the options, explaining why certain diagnoses are more or less likely. The process of elimination is clear, logical, and clinically astute, demonstrating a sophisticated understanding of disease presentation.

Level 4 (Good): The answer provides a relevant list of differential diagnoses and uses a logical process to narrow them down. The reasoning is clear and correct, though it may not explore the full spectrum of possibilities or the nuances between diagnoses as deeply as a Level 5 response.

Level 3 (Acceptable): The answer presents a limited but reasonable list of the most common differential diagnoses. It makes a plausible choice but the reasoning for excluding other options is superficial, weak, or absent. The process is functional but lacks depth.

Level 2 (Poor): The answer mentions one or two possible diagnoses but fails to create a structured list or engage in a meaningful comparison. It may jump to a conclusion prematurely or miss several obvious and important alternative diagnoses.

Level 1 (Very Poor): The answer fails to perform a differential diagnosis. It either provides a single answer with no consideration of alternatives or generates a list that is irrelevant, illogical, or factually incorrect.

\textbf{2. Safety Check}

This category assesses the ability to identify, prioritize, and mitigate potential risks to the patient. It reflects clinical responsibility and risk management.

Level 5 (Excellent): The answer demonstrates exceptional foresight. It not only identifies critical "red flag" conditions but also masterfully weighs complex, competing risks (e.g., balancing the risks of a treatment against the risks of a disease). It correctly prioritizes the most immediate or severe threat and explains its risk-benefit analysis with clinical wisdom.

Level 4 (Good): The answer actively identifies and addresses significant safety concerns. It may discuss contraindications or weigh the pros and cons of different options from a safety perspective. The reasoning is proactive and demonstrates a strong awareness of patient safety.

Level 3 (Acceptable): The answer identifies and avoids obvious, direct risks or contraindications. It follows standard safety protocols (e.g., recommends confirming a diagnosis before treatment) but does not proactively analyze more complex or subtle risks. The behavior is reactive rather than proactive.

Level 2 (Poor): The answer misses a significant safety concern or mentions a risk but fails to act on it or incorporate it into the final decision. The safety awareness is present but insufficient for safe clinical practice.

Level 1 (Very Poor): The answer makes a recommendation that is dangerous, contraindicated, or completely ignores a critical, life-threatening risk. The response poses a direct threat to patient safety.

\textbf{3. Evidence Citation}

This category assesses the ability to ground its reasoning in specific, relevant evidence, both from the patient's data and from established medical knowledge.

Level 5 (Excellent): The answer seamlessly integrates multiple pieces of evidence (e.g., symptoms, lab values, patient history, and pharmacological data) into a cohesive and compelling argument. It not only cites evidence but also explains the significance and weight of key findings, demonstrating how specific evidence shifts the diagnostic probabilities. The reasoning is transparent and deeply rooted in evidence-based principles.

Level 4 (Good): The answer consistently and accurately cites relevant evidence to support its main arguments and conclusions. It clearly links specific findings ("Because of X...") to its reasoning ("...we can conclude Y."). It effectively uses a combination of patient-specific data and general medical facts.

Level 3 (Acceptable): The answer cites the most obvious pieces of evidence to support its final conclusion but may ignore other relevant data. The link between evidence and conclusion is present but may be simplistic. The reasoning is supported, but not robustly.

Level 2 (Poor): The answer mentions pieces of evidence from the prompt but fails to logically connect them to its reasoning or conclusion. The citation feels like a simple restatement of facts rather than an integrated part of an argument.

Level 1 (Very Poor): The answer makes claims without any supporting evidence, misinterprets the provided evidence, or uses irrelevant information to justify its conclusions.

Evaluation Output Template:

Please use the following format for your response.

\begin{itemize}
    \item DDx Score: [1-5]
    \item Safety Check Score: [1-5]
    \item Evidence Citation Score: [1-5]
    \item Rationale for DDx: [Provide a brief, specific justification for the score, citing examples from the  response.]
    \item Rationale for Safety Check: [Provide a brief, specific justification for the score, citing examples from the  response.]
    \item Rationale for Evidence Citing: [Provide a brief, specific justification for the score, citing examples from the  response.]
\end{itemize}

Here are the Question and Answer:

[QUESTION]

\end{tcolorbox}

\begin{figure*}
    \centering
    \includegraphics[width=0.7\linewidth]{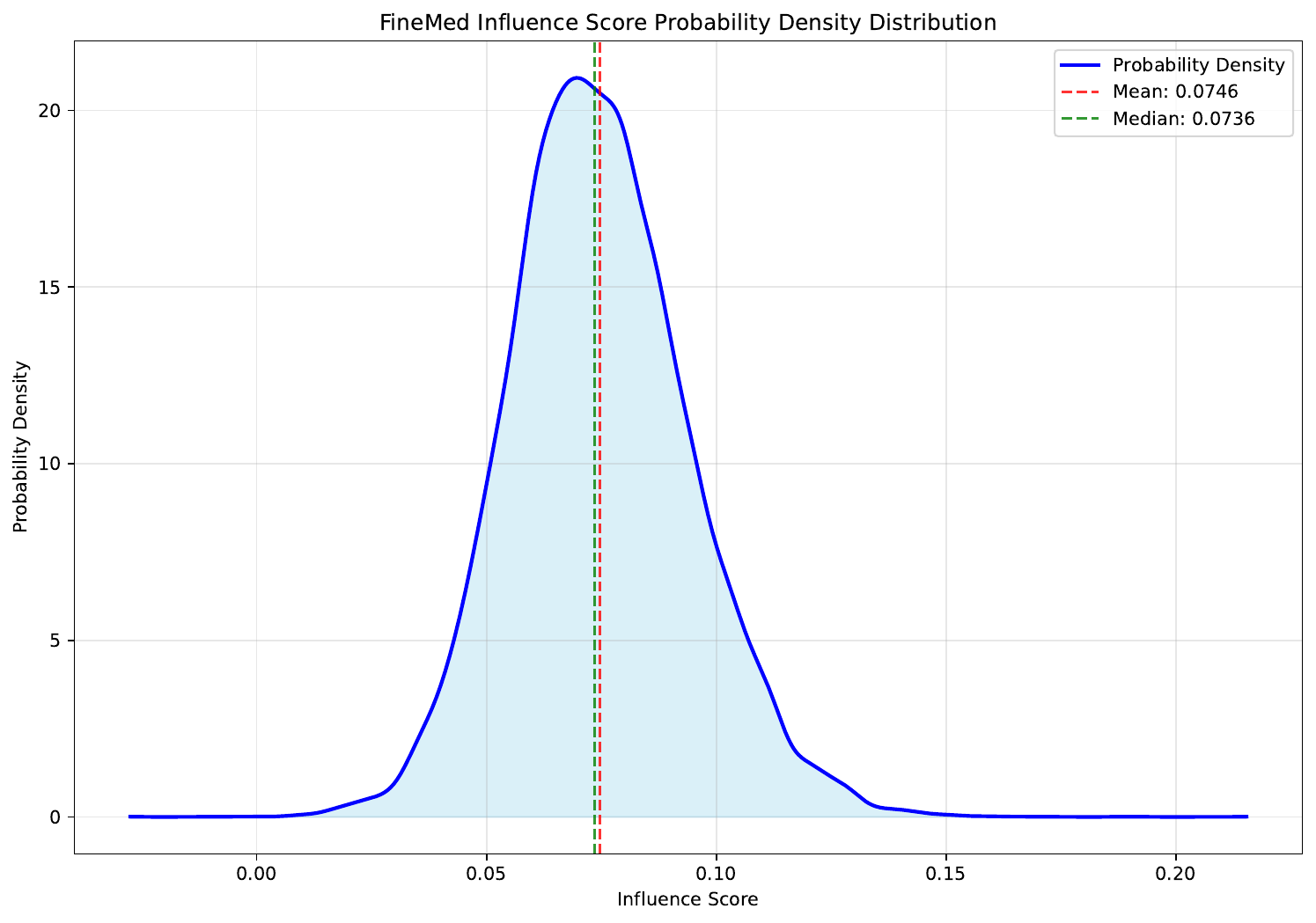}
    \caption{Influence score distribution of FineMed.}
    \label{fig:inf_finemed}
\end{figure*}

\begin{figure*}
    \centering
    \includegraphics[width=0.7\linewidth]{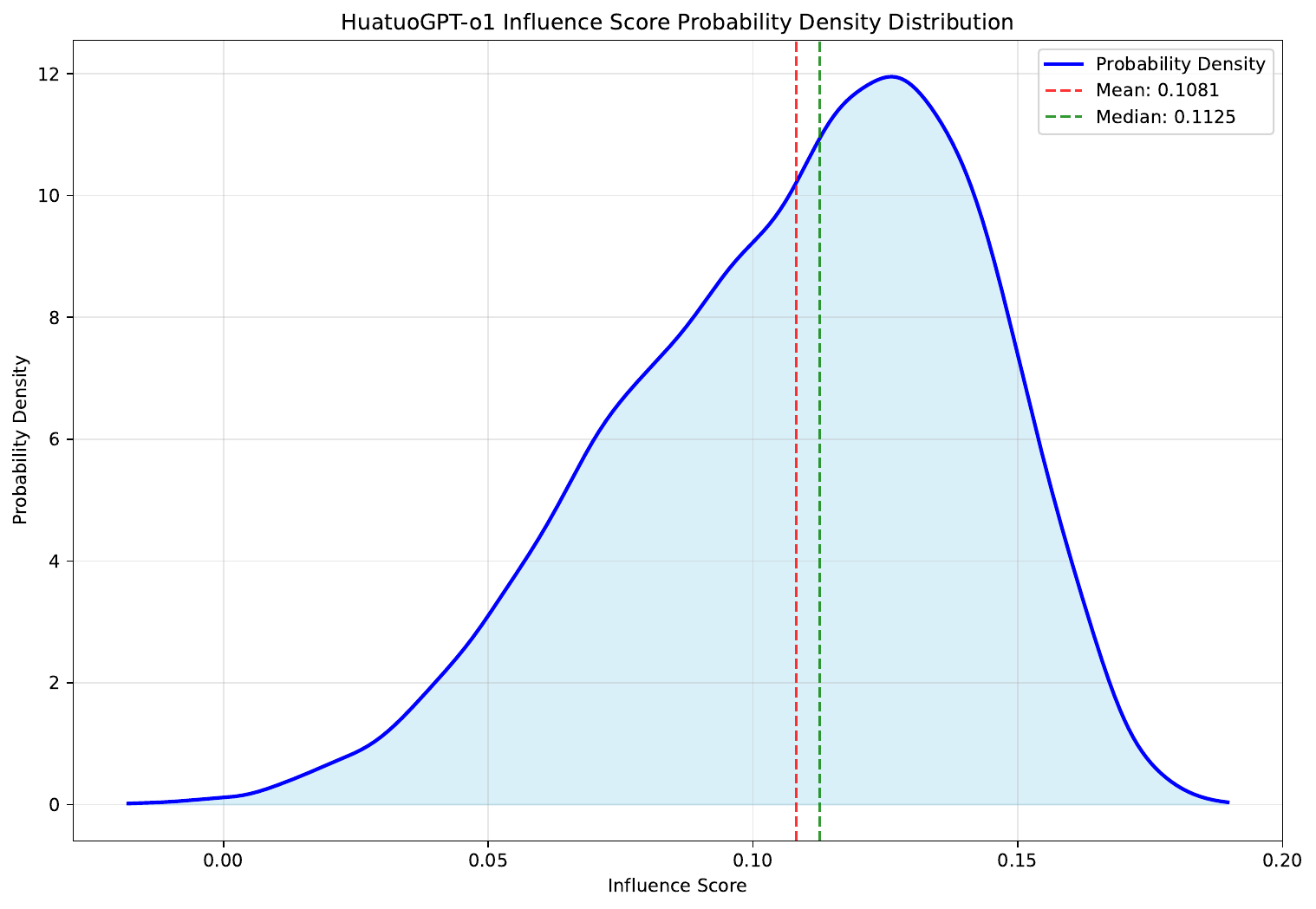}
    \caption{Influence score distribution of Huatuo.}
    \label{fig:inf_huatuo}
\end{figure*}

\begin{figure*}
    \centering
    \includegraphics[width=0.7\linewidth]{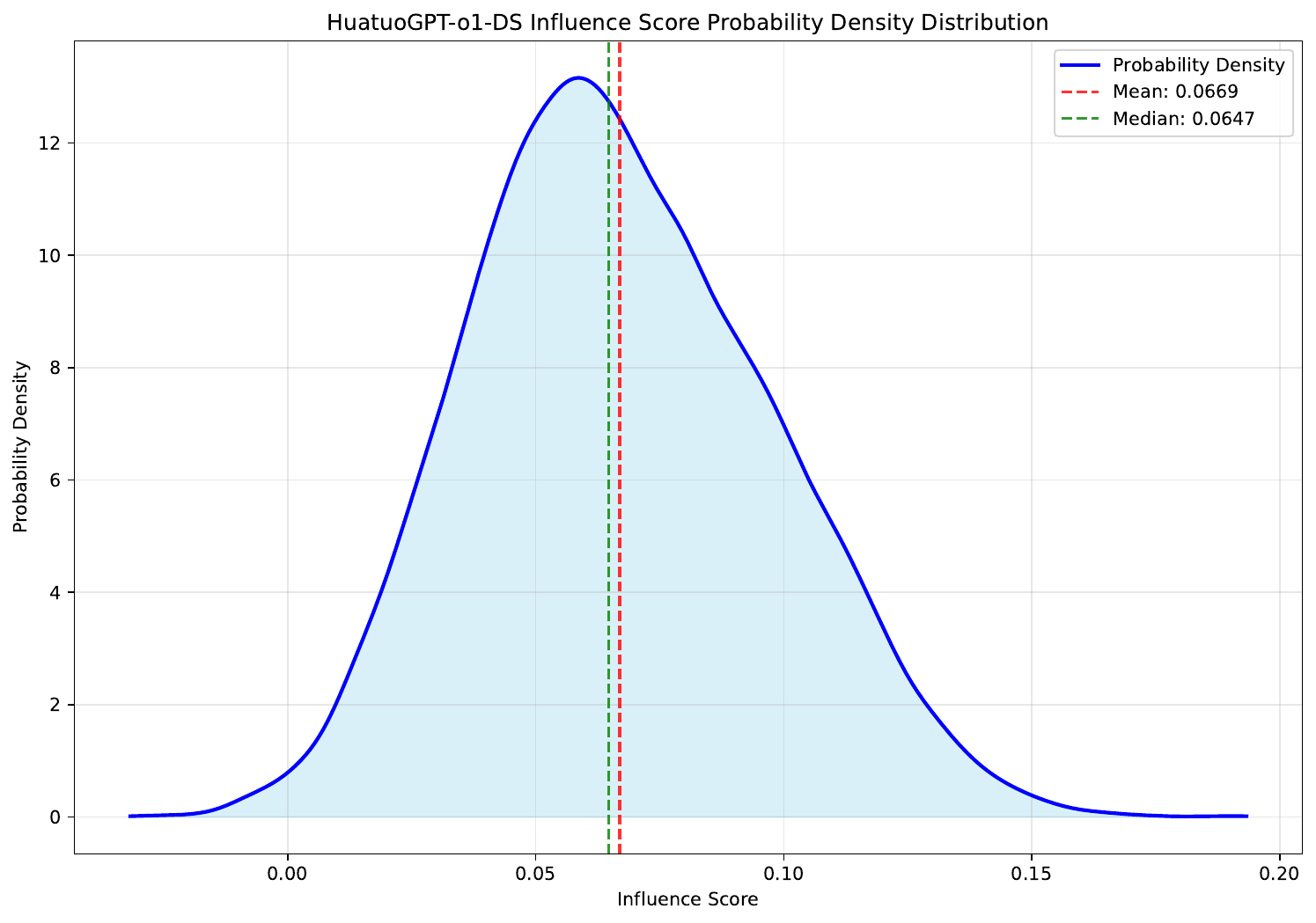}
    \caption{Influence score distribution of Huatuo-DS.}
    \label{fig:inf_huatuods}
\end{figure*}

\begin{figure*}
    \centering
    \includegraphics[width=0.7\linewidth]{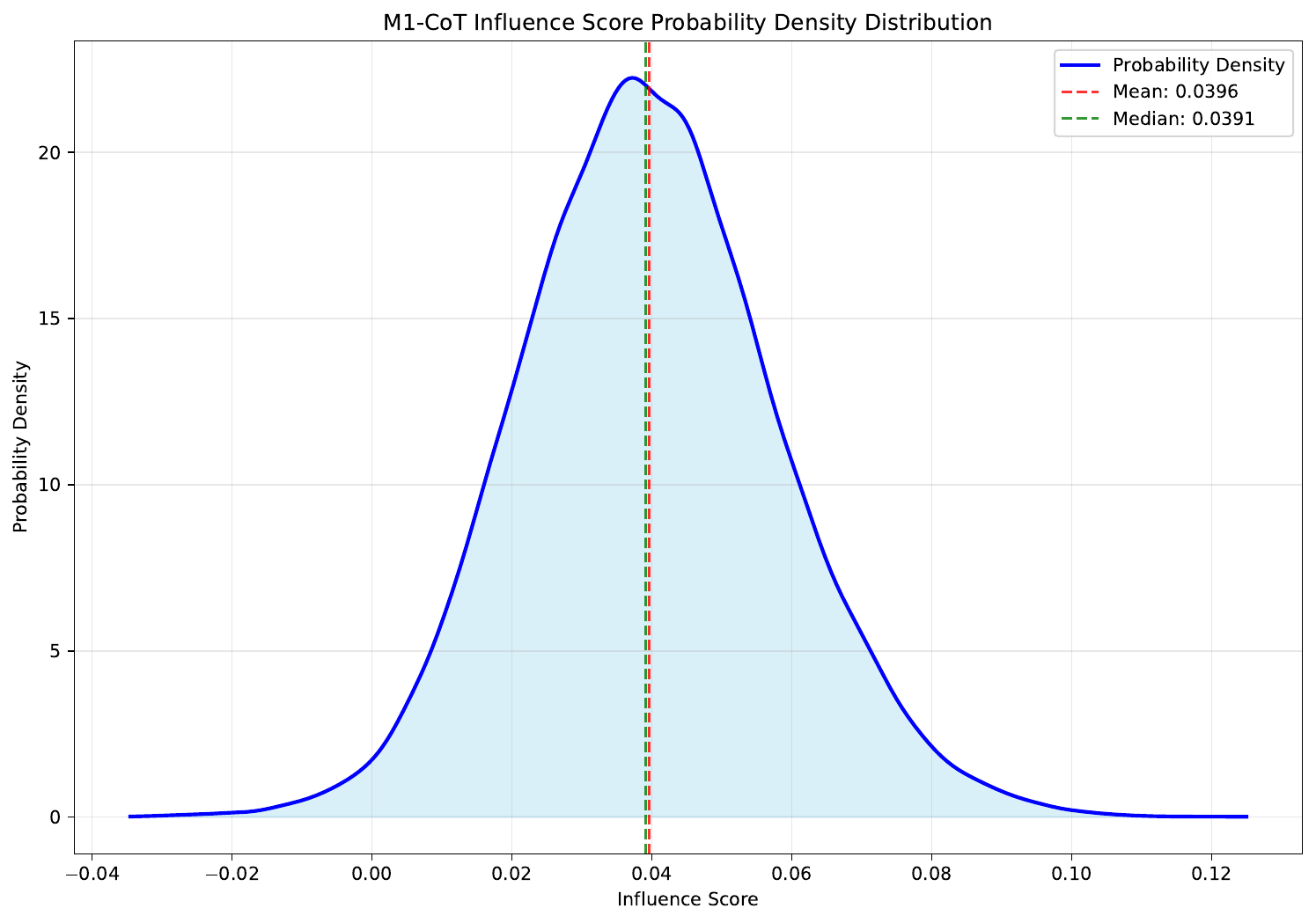}
    \caption{Influence score distribution of m1.}
    \label{fig:inf_m1}
\end{figure*}

\begin{figure*}
    \centering
    \includegraphics[width=0.7\linewidth]{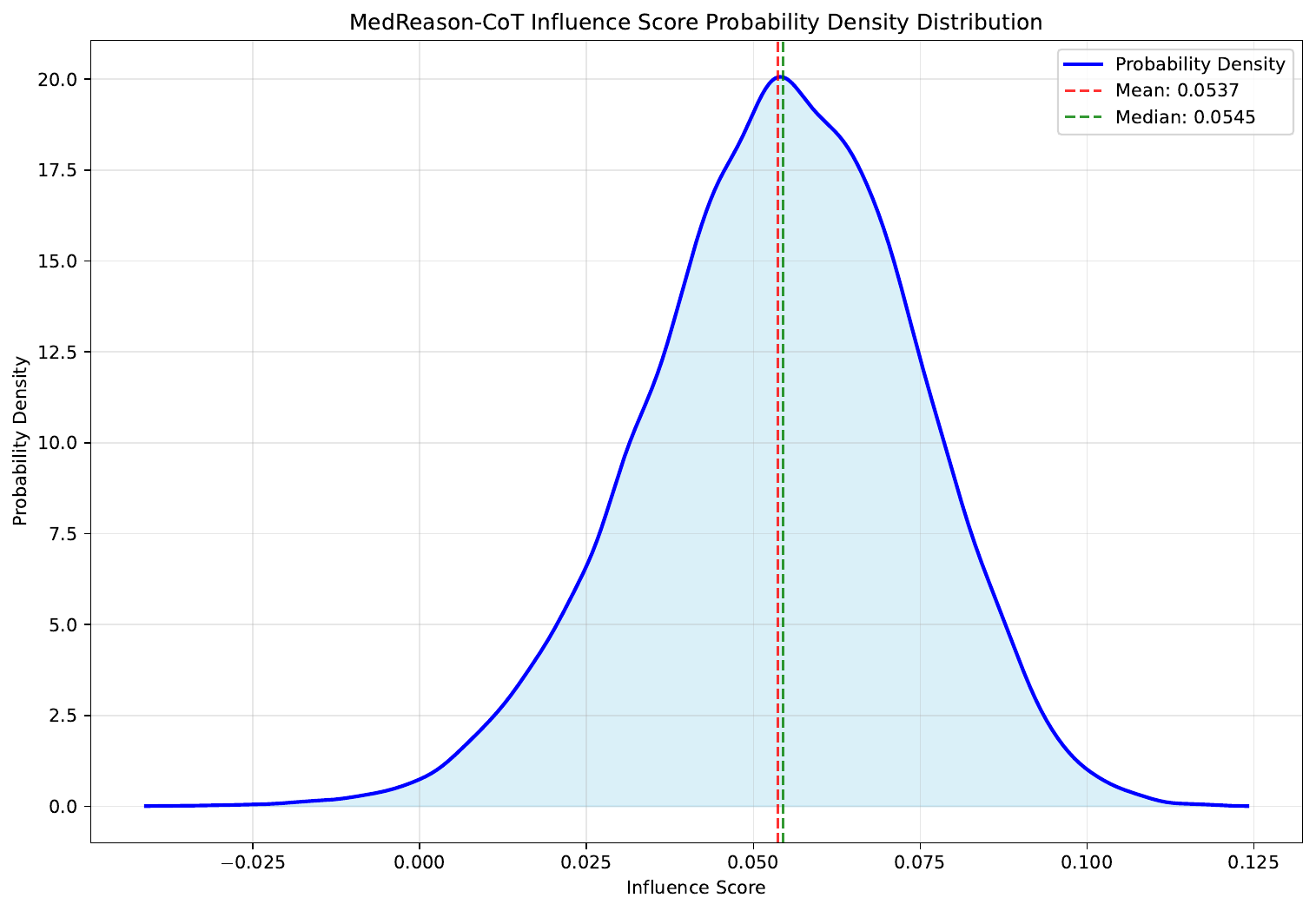}
    \caption{Influence score distribution of MedReason.}
    \label{fig:inf_medreason}
\end{figure*}

\section{Details of Efficiency Analysis}

\label{app:cost}
We use Eq. \ref{eq:train} to approximate FLOPs for training on transformer-style models.
\begin{equation}
    \label{eq:train}
    F_\text{train} = 6 \times L \times \ H^2 \times T \times |D_\text{train}| \times E 
\end{equation}
where $L$ denotes the number of model layers, $H$ denotes the hidden size, $T$ denotes number of tokens per sample, $|D_\text{train}|$ denotes the number of training samples, and $E$ denotes the number of training epochs.
Similarly, the inference FLOPs can be approximated as:
\begin{equation}
    \label{eq:infer}
    F_\text{infer} = 2  \times L \times \ H^2 \times T \times |D_\text{infer}|
\end{equation}
where $|D_\text{infer}|$ denotes the number of samples to infer on.
For LoRA fine-tuning, the formula can be adapted to account for the reduced number of trainable parameters. 
The core is replacing the quadratic dependency on the hidden size ($H^2$) with a term proportional to the LoRA rank ($r$) of the decomposition:
\begin{equation}
\label{eq:lora}
F_\text{LoRA} = 12 \times k \times L \times H \times r \times T \times |D_\text{train}| \times E
\end{equation}
where $r$ denotes the rank of the two LoRA matrices, and $k$ denotes the number of matrices adapted with LoRA per layer ($k=3$ in our experiment  since LoRA is applied to query, key and value matrices in the self-attention blocks).

\section{Full Experimental Results}
\label{app:full_all}

\subsection{Main Results}
\label{app:full_main_results}

\begin{table*}
\small
\centering
\begin{tabular}{@{}lcccccccccccc@{}}
\toprule
\textbf{Model} &
  \multicolumn{1}{l}{\textbf{MedQ}} &
  \multicolumn{1}{l}{\textbf{MedM}} &
  \multicolumn{1}{l}{\textbf{MMLU}} &
  \multicolumn{1}{l}{\textbf{Avg$_S$}} &
  \multicolumn{1}{l}{\textbf{HLE}} &
  \multicolumn{1}{l}{\textbf{MeB4}} &
  \multicolumn{1}{l}{\textbf{MeB5}} &
  \multicolumn{1}{l}{\textbf{MedX}} &
  \multicolumn{1}{l}{\textbf{MedG}} &
  \multicolumn{1}{l}{\textbf{MetM}} &
  \multicolumn{1}{l}{\textbf{Avg$_C$}} &
  \multicolumn{1}{l}{\textbf{Avg$_A$}} \\ \midrule
GPT-4.1          & 84.29 & 73.34 & 82.46 & 80.03 & 7.77  & 71.75 & 70.13 & 42.00 & 64.44 & 70.79 & 54.48 & 63.00 \\
DeepSeek-V3-0324 & 73.76 & 55.10 & 62.90 & 63.92 & 6.80  & 73.38 & 66.56 & 38.04 & 59.77 & 65.84 & 51.73 & 55.79 \\
Gemini-2.5-flash & 90.73 & 77.34 & 90.36 & 86.14 & 11.65 & 82.14 & 76.62 & 36.82 & 61.55 & 77.13 & 57.65 & 67.15 \\ \midrule
DeepSeek-R1-0528 & 92.85 & 76.55 & 91.28 & 86.89 & 13.59 & 83.44 & 54.22 & 38.61 & 59.88 & 72.91 & 53.78 & 64.81 \\
QwQ-32B          & 75.10 & 63.45 & 78.97 & 72.51 & 12.62 & 67.86 & 59.09 & 22.65 & 48.44 & 63.80 & 45.74 & 54.66 \\
o4-mini-medium   & 64.73 & 61.44 & 81.08 & 69.08 & 13.59 & 70.78 & 71.10 & 40.78 & 60.46 & 76.11 & 55.47 & 60.01 \\
Gemini-2.5-pro   & 78.00 & 79.75 & 85.67 & 81.14 & 15.53 & 84.42 & 78.57 & 42.37 & 62.11 & 73.05 & 59.34 & 66.61 \\ \bottomrule
\end{tabular}
\caption{Full downstream task results of general reasoning and non-reasoning models.}
\label{tab:api_model}
\end{table*}

\begin{table*}
\small
\centering
\begin{tabular}{@{}lcccccccccccc@{}}
\toprule
\textbf{Model} &
  \multicolumn{1}{l}{\textbf{MedQ}} &
  \multicolumn{1}{l}{\textbf{MedM}} &
  \multicolumn{1}{l}{\textbf{MMLU}} &
  \multicolumn{1}{l}{\textbf{Avg$_S$}} &
  \multicolumn{1}{l}{\textbf{HLE}} &
  \multicolumn{1}{l}{\textbf{MeB4}} &
  \multicolumn{1}{l}{\textbf{MeB5}} &
  \multicolumn{1}{l}{\textbf{MedX}} &
  \multicolumn{1}{l}{\textbf{MedG}} &
  \multicolumn{1}{l}{\textbf{MetM}} &
  \multicolumn{1}{l}{\textbf{Avg$_C$}} &
  \multicolumn{1}{l}{\textbf{Avg$_A$}} \\ \midrule
\textbf{Llama3.1-8B-Inst} & 53.26 & 53.15 & 61.57 & 55.99 & 11.65 & 37.99 & 36.04 & 15.63 & 42.26 & 37.22 & 30.13 & 38.75 \\ \midrule
\textbf{Huatuo}           & 58.68 & 47.79 & 57.85 & 54.77 & 24.27 & 44.16 & 40.91 & 20.33 & 43.28 & 53.68 & 37.77 & 43.44 \\
50\% Random               & 57.74 & 48.39 & 57.94 & 54.69 & 23.30 & 46.75 & 40.26 & 18.49 & 41.76 & 35.54 & 34.35 & 41.13 \\
50\% DIQ                  & 57.50 & 49.34 & 59.32 & 55.39 & 22.33 & 41.88 & 39.61 & 23.14 & 41.50 & 43.48 & 35.32 & 42.01 \\ \midrule
10\% Random               & 52.87 & 47.33 & 56.75 & 52.32 & 15.53 & 39.94 & 30.84 & 17.51 & 40.71 & 35.32 & 29.98 & 37.42 \\
10\% DIQ                  & 58.13 & 53.57 & 62.63 & 58.11 & 25.24 & 44.48 & 40.40 & 17.59 & 43.38 & 50.91 & 37.00 & 44.04 \\ \midrule
1\% Random                & 54.75 & 43.99 & 55.19 & 51.31 & 15.86 & 42.50 & 37.71 & 13.63 & 44.75 & 46.39 & 33.47 & 39.42 \\
1\% DIQ                   & 56.64 & 50.16 & 62.81 & 56.54 & 13.59 & 47.40 & 47.75 & 14.45 & 45.86 & 46.39 & 35.91 & 42.78 \\ \midrule
\textbf{Huatuo-DS}        & 67.24 & 58.26 & 74.38 & 66.63 & 11.65 & 57.47 & 51.62 & 15.71 & 43.01 & 53.68 & 38.86 & 48.11 \\
50\% Random               & 64.34 & 55.77 & 71.72 & 63.94 & 7.77  & 53.90 & 51.62 & 15.59 & 42.71 & 53.82 & 37.57 & 46.36 \\
50\% DIQ                  & 66.46 & 56.49 & 72.18 & 65.04 & 12.62 & 54.22 & 50.65 & 16.45 & 42.78 & 53.17 & 38.32 & 47.22 \\ \midrule
10\% Random               & 65.12 & 57.38 & 71.26 & 64.59 & 12.62 & 57.14 & 48.05 & 14.41 & 41.65 & 52.08 & 37.66 & 46.63 \\
10\% DIQ                  & 67.87 & 57.57 & 73.00 & 66.15 & 14.56 & 59.35 & 50.97 & 15.55 & 44.89 & 54.84 & 40.03 & 48.73 \\ \midrule
1\% Random                & 56.95 & 49.63 & 62.53 & 56.37 & 7.77  & 45.86 & 41.88 & 13.18 & 40.29 & 41.66 & 31.77 & 39.97 \\
1\% DIQ                   & 63.16 & 53.93 & 63.36 & 60.15 & 16.50 & 46.10 & 44.48 & 25.47 & 39.27 & 32.19 & 37.96 & 45.36 \\ \midrule
\textbf{FineMed}          & 40.22 & 51.26 & 51.61 & 47.70 & 16.50 & 46.10 & 44.48 & 25.47 & 39.27 & 32.19 & 34.00 & 38.57 \\
50\% Random               & 40.38 & 33.83 & 37.47 & 37.23 & 17.48 & 42.86 & 43.83 & 21.43 & 40.12 & 38.53 & 34.04 & 35.10 \\
50\% DIQ                  & 42.66 & 35.24 & 39.49 & 39.13 & 18.45 & 50.65 & 43.83 & 19.47 & 41.87 & 37.14 & 35.24 & 36.53 \\ \midrule
10\% Random               & 51.14 & 39.04 & 45.27 & 45.15 & 16.50 & 45.13 & 42.50 & 16.49 & 42.89 & 40.93 & 34.07 & 37.77 \\
10\% DIQ                  & 51.61 & 40.40 & 45.91 & 45.97 & 17.48 & 48.05 & 43.83 & 18.57 & 44.87 & 43.55 & 36.06 & 39.36 \\ \midrule
1\% Random                & 51.61 & 48.98 & 58.68 & 53.09 & 11.65 & 45.45 & 42.86 & 13.59 & 40.29 & 35.76 & 31.60 & 38.76 \\
1\% DIQ                   & 53.50 & 54.15 & 66.76 & 58.14 & 12.62 & 45.45 & 42.21 & 13.80 & 44.28 & 40.35 & 33.12 & 41.46 \\ \midrule
\textbf{m1}               & 75.88 & 64.33 & 82.83 & 74.35 & 16.50 & 66.56 & 60.06 & 17.35 & 43.68 & 58.99 & 43.86 & 54.02 \\
50\% Random               & 74.31 & 62.68 & 82.46 & 73.15 & 14.56 & 64.29 & 58.44 & 19.39 & 41.64 & 61.62 & 43.32 & 53.27 \\
50\% DIQ                  & 74.86 & 63.47 & 82.19 & 73.51 & 17.48 & 64.61 & 56.82 & 17.76 & 43.35 & 61.54 & 43.59 & 53.56 \\ \midrule
10\% Random               & 75.26 & 60.53 & 79.98 & 71.81 & 12.62 & 60.39 & 57.79 & 17.59 & 46.97 & 60.52 & 42.65 & 52.41 \\
10\% DIQ                  & 74.39 & 61.25 & 79.80 & 71.92 & 16.50 & 62.66 & 58.44 & 17.59 & 46.61 & 62.49 & 44.05 & 53.30 \\ \midrule
1\% Random                & 71.09 & 58.52 & 77.78 & 69.13 & 4.85  & 55.84 & 53.25 & 16.41 & 46.50 & 58.49 & 39.22 & 49.19 \\
1\% DIQ                   & 72.03 & 60.51 & 76.95 & 69.83 & 10.68 & 61.36 & 54.87 & 16.82 & 46.31 & 59.72 & 41.63 & 51.03 \\ \midrule
\textbf{MedReason}        & 61.51 & 42.34 & 73.28 & 59.04 & 9.71  & 34.87 & 36.17 & 15.96 & 43.38 & 37.22 & 29.55 & 39.38 \\
50\% Random               & 60.41 & 50.39 & 66.48 & 59.09 & 11.65 & 36.69 & 32.14 & 18.29 & 48.92 & 34.01 & 30.28 & 39.89 \\
50\% DIQ                  & 63.24 & 53.07 & 71.07 & 62.46 & 16.50 & 36.69 & 28.90 & 18.29 & 48.48 & 35.40 & 30.71 & 41.29 \\ \midrule
10\% Random               & 58.99 & 51.61 & 64.46 & 58.35 & 11.65 & 33.44 & 26.30 & 15.84 & 42.27 & 44.57 & 29.01 & 38.79 \\
10\% DIQ                  & 59.62 & 49.18 & 59.41 & 56.07 & 22.33 & 44.81 & 44.48 & 17.47 & 39.74 & 48.94 & 36.30 & 42.89 \\ \midrule
1\% Random                & 45.40 & 35.38 & 41.41 & 40.73 & 18.45 & 40.58 & 37.99 & 15.35 & 42.74 & 46.39 & 33.58 & 35.97 \\
1\% DIQ                   & 46.11 & 44.59 & 51.61 & 47.44 & 21.36 & 41.23 & 38.31 & 17.80 & 48.48 & 46.39 & 35.60 & 39.54 \\ \midrule
\textbf{UltraMedical}     & 67.24 & 51.95 & 69.70 & 62.96 & 23.30 & 51.95 & 50.32 & 15.22 & 43.63 & 45.45 & 38.31 & 46.53 \\
50\% Random               & 67.09 & 49.59 & 67.31 & 61.33 & 14.56 & 52.57 & 46.43 & 13.18 & 44.60 & 42.83 & 35.70 & 44.24 \\ 
50\% DIQ                  & 67.09 & 53.50 & 69.42 & 63.34 & 18.45 & 53.25 & 51.30 & 13.80 & 44.95 & 43.92 & 37.61 & 46.19 \\ \midrule
10\% Random               & 66.22 & 55.73 & 71.35 & 64.43 & 10.68 & 56.49 & 46.78 & 13.80 & 48.02 & 43.70 & 36.58 & 45.86 \\
10\% DIQ                  & 67.79 & 54.41 & 72.82 & 65.01 & 12.62 & 56.57 & 47.40 & 15.18 & 48.24 & 44.28 & 37.38 & 46.59 \\ \midrule
1\% Random                & 65.28 & 54.77 & 71.72 & 63.92 & 12.62 & 53.90 & 45.78 & 13.88 & 47.40 & 43.26 & 36.14 & 45.40 \\
1\% DIQ                   & 66.69 & 58.52 & 71.63 & 65.61 & 11.65 & 57.14 & 46.75 & 13.59 & 47.21 & 44.65 & 36.83 & 46.43 \\ \bottomrule
\end{tabular}
\caption{Full downstream task results of trained Llama-3.1-8B-Instruct using full dataset, random subset, and our DIQ selected data at 1\%, 10\%, and 50\% data keeping ratios.}
\label{tab:llama31_8b}
\end{table*}

Full downstream task results of general non-reasoning and reasoning models, and Llama-3.1-8B-Instruct trained on different selection setting data are provided in Tables \ref{tab:api_model} and  \ref{tab:llama31_8b}.

\subsection{QA Experiment}
\label{app:full_qa}
Full downstream task results of QA experiment are provided in Table \ref{tab:full_qa}.

\begin{table*}[]
\small
\centering
\begin{tabular}{@{}lcccccccccccc@{}}
\toprule
\textbf{Setting}         & \multicolumn{1}{l}{\textbf{MedQ}} & \multicolumn{1}{l}{\textbf{MedM}} & \multicolumn{1}{l}{\textbf{MMLU}} & \multicolumn{1}{l}{\textbf{Avg$_S$}} & \multicolumn{1}{l}{\textbf{HLE}} & \multicolumn{1}{l}{\textbf{MeB4}} & \multicolumn{1}{l}{\textbf{MeB5}} & \multicolumn{1}{l}{\textbf{MedX}} & \multicolumn{1}{l}{\textbf{MedG}} & \multicolumn{1}{l}{\textbf{MetM}} & \multicolumn{1}{l}{\textbf{Avg$_C$}} & \multicolumn{1}{l}{\textbf{Avg$_A$}} \\ \midrule
\multicolumn{13}{c}{\textbf{Llama3.1-8B-Instruct}}                                                                             \\ \midrule
\textbf{m1-23k-QA}       &                                   &                                   &                                   &                                      &                                  &                                   &                                   &                                   &                                   &                                   &                                      &                                      \\
50\% Random              & 67.64                             & 58.40                             & 75.85                             & 67.30                                & 8.74                             & 57.47                             & 52.27                             & 14.90                             & 46.65                             & 53.90                             & 38.99                                & 48.42                                \\
50\% DIQ                 & 66.93                             & 58.57                             & 75.85                             & 67.12                                & 19.42                            & 54.22                             & 52.92                             & 13.80                             & 48.99                             & 55.35                             & 40.78                                & 49.56                                \\ \midrule
10\% Random              & 65.91                             & 58.36                             & 75.67                             & 66.65                                & 10.68                            & 55.84                             & 48.38                             & 13.63                             & 46.88                             & 54.19                             & 38.27                                & 47.73                                \\
10\% DIQ                 & 66.06                             & 57.73                             & 75.21                             & 66.33                                & 20.39                            & 55.52                             & 50.32                             & 14.08                             & 46.91                             & 54.84                             & 40.34                                & 49.01                                \\ \midrule
1\% Random               & 60.09                             & 51.11                             & 64.83                             & 58.68                                & 12.62                            & 47.08                             & 43.51                             & 18.16                             & 49.15                             & 46.25                             & 36.13                                & 43.64                                \\
1\% DIQ                  & 62.69                             & 53.41                             & 64.92                             & 60.34                                & 16.50                            & 52.27                             & 46.43                             & 16.73                             & 48.53                             & 45.88                             & 37.72                                & 45.26                                \\ \midrule
\textbf{MedReason-QA}    &                                   &                                   &                                   &                                      &                                  &                                   &                                   &                                   &                                   &                                   &                                      &                                      \\
50\% Random              & 58.99                             & 51.49                             & 72.73                             & 61.07                                & 20.39                            & 36.36                             & 33.12                             & 17.14                             & 48.84                             & 38.75                             & 32.43                                & 41.98                                \\
50\% DIQ                 & 59.62                             & 54.58                             & 75.76                             & 63.32                                & 20.39                            & 45.45                             & 38.96                             & 23.35                             & 47.52                             & 44.94                             & 36.77                                & 45.62                                \\ \midrule
10\% Random              & 61.19                             & 50.75                             & 67.13                             & 59.69                                & 17.48                            & 37.66                             & 34.74                             & 17.96                             & 48.81                             & 40.93                             & 32.93                                & 41.85                                \\
10\% DIQ                 & 60.64                             & 54.27                             & 72.54                             & 62.48                                & 29.13                            & 53.90                             & 49.35                             & 18.65                             & 39.76                             & 48.22                             & 39.84                                & 47.38                                \\ \midrule
1\% Random               & 44.62                             & 36.53                             & 46.19                             & 42.45                                & 24.27                            & 43.51                             & 39.61                             & 14.57                             & 46.17                             & 38.24                             & 34.40                                & 37.08                                \\
1\% DIQ                  & 45.40                             & 42.96                             & 53.08                             & 47.15                                & 26.21                            & 43.18                             & 39.94                             & 17.18                             & 47.04                             & 46.54                             & 36.68                                & 40.17                                \\ \midrule
\multicolumn{13}{c}{\textbf{Qwen3-8B}}                                                                             \\ \midrule
\textbf{m1-23k-QA}       &                                   &                                   &                                   &                                      &                                  &                                   &                                   &                                   &                                   &                                   &                                      &                                      \\
50\% Random              & 64.41                             & 55.87                             & 78.70                              & 66.33                                & 13.59                            & 52.60                             & 44.81                             & 15.59                             & 45.28                             & 51.64                             & 37.25                                & 46.94                                \\
50\% DIQ                 & 65.75                             & 55.82                             & 77.87                             & 66.48                                & 18.45                            & 57.14                             & 46.75                             & 16.86                             & 47.92                             & 50.47                             & 39.60                                & 48.56                                \\ \midrule
10\% Random              & 64.02                             & 57.85                             & 78.79                             & 66.89                                & 13.59                            & 52.92                             & 43.51                             & 15.39                             & 48.06                             & 52.44                             & 37.65                                & 47.40                                \\
10\% DIQ                 & 65.75                             & 57.85                             & 79.71                             & 67.77                                & 17.48                            & 49.03                             & 46.75                             & 15.22                             & 49.06                             & 53.02                             & 38.43                                & 48.21                                \\ \midrule
1\% Random               & 77.14                             & 63.54                             & 84.11                             & 74.93                                & 16.50                            & 64.94                             & 57.47                             & 15.22                             & 54.45                             & 61.76                             & 45.06                                & 55.01                                \\
1\% DIQ                  & 77.14                             & 64.26                             & 85.40                             & 75.60                                & 20.39                            & 65.91                             & 61.69                             & 23.35                             & 54.96                             & 62.56                             & 48.14                                & 57.30                                \\ \midrule
\textbf{MedReason-QA}    &                                   &                                   &                                   &                                      &                                  &                                   &                                   &                                   &                                   &                                   &                                      &                                      \\
50\% Random              & 56.32                             & 51.90                             & 75.57                             & 61.26                                & 19.42                            & 41.56                             & 33.12                             & 16.29                             & 50.16                             & 37.58                             & 33.02                                & 42.44                                \\
50\% DIQ                 & 58.29                             & 53.62                             & 74.84                             & 62.25                                & 25.24                            & 40.58                             & 42.86                             & 16.69                             & 48.44                             & 40.35                             & 35.69                                & 44.55                                \\ \midrule
10\% Random              & 52.40                             & 50.94                             & 66.76                             & 56.70                                & 19.42                            & 40.91                             & 35.06                             & 14.73                             & 47.57                             & 28.91                             & 31.10                                & 39.63                                \\
10\% DIQ                 & 56.64                             & 52.98                             & 69.88                             & 59.83                                & 19.42                            & 49.03                             & 45.45                             & 17.96                             & 48.47                             & 51.27                             & 38.60                                & 45.68                                \\ \midrule
1\% Random               & 35.19                             & 41.55                             & 53.81                             & 43.52                                & 7.77                             & 39.94                             & 30.52                             & 13.02                             & 29.87                             & 35.83                             & 26.16                                & 31.94                                \\
1\% DIQ                  & 52.95                             & 46.98                             & 64.46                             & 54.80                                & 14.56                            & 44.16                             & 41.56                             & 14.12                             & 43.27                             & 45.96                             & 33.94                                & 40.89                                \\ \bottomrule
\end{tabular}
\caption{Full downstream task results of Llama3.1-8B-Instruct and Qwen3-8B under training on 1\%, 10\%, and 50\% randomly selected and DIQ-selected QA datasets.}
\label{tab:full_qa}
\end{table*}

\subsection{Ablation Study Experiment}
\label{app:full_ablation}

Full downstream task results of DIQ ablation study are provided in Table \ref{tab:full_ablation}. 

\begin{table*}
\centering
\small
\begin{tabular}{@{}lcccccccccccc@{}}
\toprule
\textbf{Setting} &
  \multicolumn{1}{l}{\textbf{MedQ}} &
  \multicolumn{1}{l}{\textbf{MedM}} &
  \multicolumn{1}{l}{\textbf{MMLU}} &
  \multicolumn{1}{l}{\textbf{Avg$_S$}} &
  \multicolumn{1}{l}{\textbf{HLE}} &
  \multicolumn{1}{l}{\textbf{MeB4}} &
  \multicolumn{1}{l}{\textbf{MeB5}} &
  \multicolumn{1}{l}{\textbf{MedX}} &
  \multicolumn{1}{l}{\textbf{MedG}} &
  \multicolumn{1}{l}{\textbf{MetM}} &
  \multicolumn{1}{l}{\textbf{Avg$_C$}} &
  \multicolumn{1}{l}{\textbf{Avg$_A$}} \\ \midrule
\multicolumn{13}{c}{\textbf{Llama3.1-8B-Instruct}}                                                                             \\ \midrule
50\% Influence                         & 56.95 & 50.11 & 56.75 & 54.60 & 22.33 & 45.45 & 36.36 & 21.96 & 41.72 & 41.30 & 34.85 & 41.44 \\
50\% Overall                           & 58.21 & 48.86 & 56.75 & 54.61 & 22.33 & 44.16 & 37.66 & 17.76 & 44.33 & 35.69 & 33.66 & 40.64 \\
50\% Knowledge                         & 57.97 & 48.12 & 60.51 & 55.53 & 26.21 & 43.51 & 36.69 & 18.33 & 41.02 & 36.78 & 33.76 & 41.02 \\
50\% Reasoning                         & 57.11 & 47.43 & 56.20 & 53.58 & 21.36 & 51.30 & 38.31 & 18.29 & 43.57 & 34.89 & 34.62 & 40.94 \\
50\% DIQ & 57.50 & 49.34 & 59.32 & 55.39 & 22.33 & 41.88 & 39.61 & 23.14 & 41.50 & 43.48 & 35.32 & \textbf{42.01} \\ \midrule
10\% Influence                         & 58.37 & 51.78 & 61.98 & 57.38 & 25.24 & 40.26 & 38.31 & 20.20 & 42.35 & 45.01 & 35.23 & 42.61 \\
10\% Overall                           & 54.36 & 51.30 & 59.69 & 55.12 & 23.30 & 46.10 & 40.26 & 20.24 & 42.25 & 50.91 & 37.18 & 43.16 \\
10\% Knowledge                         & 53.42 & 49.37 & 56.38 & 53.06 & 23.30 & 43.18 & 35.06 & 18.86 & 40.79 & 39.69 & 33.48 & 40.01 \\
10\% Reasoning                         & 55.15 & 50.63 & 62.35 & 56.04 & 25.24 & 42.53 & 37.66 & 17.76 & 42.17 & 43.48 & 34.81 & 41.89 \\
10\% DIQ                        & 58.13 & 53.57 & 62.63 & 58.11 & 25.24 & 44.48 & 40.40 & 17.59 & 43.38 & 50.91 & 37.00 & \textbf{44.04} \\ \midrule
1\% Influence                          & 55.85 & 46.71 & 62.73 & 55.10 & 12.62 & 51.82 & 47.73 & 13.18 & 45.17 & 48.22 & 36.46 & 42.67 \\
1\% Overall                            & 53.10 & 42.46 & 57.94 & 51.17 & 16.50 & 47.73 & 47.73 & 13.14 & 45.75 & 47.92 & 36.46 & 41.36 \\
1\% Knowledge                          & 54.20 & 44.97 & 57.30 & 52.16 & 18.45 & 51.62 & 42.21 & 12.69 & 44.95 & 47.92 & 36.31 & 41.59 \\
1\% Reasoning                          & 55.70 & 44.63 & 58.68 & 53.00 & 7.77  & 50.65 & 42.86 & 13.51 & 44.79 & 47.71 & 34.55 & 40.70 \\
1\% DIQ                         & 56.64 & 50.16 & 62.81 & 56.54 & 13.59 & 47.40 & 47.75 & 14.45 & 45.86 & 46.39 & 35.91 & \textbf{42.78} \\ \midrule
\multicolumn{13}{c}{\textbf{Qwen3-8B}}                                                                             \\ \midrule
50\% Influence                         & 60.33 & 53.86 & 74.38 & 62.86 & 20.39 & 45.13 & 44.48 & 32.78 & 45.28 & 44.72 & 38.80 & 46.82 \\
50\% Overall                           & 59.39 & 53.05 & 74.10 & 62.18 & 16.50 & 46.10 & 41.23 & 31.96 & 45.31 & 41.08 & 37.03 & 45.41 \\
50\% Knowledge                         & 59.78 & 54.46 & 73.00 & 62.41 & 13.59 & 41.88 & 39.61 & 29.76 & 44.78 & 40.50 & 35.02 & 44.15 \\
50\% Reasoning                         & 61.82 & 53.74 & 73.09 & 62.88 & 21.36 & 46.75 & 39.61 & 30.61 & 44.02 & 40.79 & 37.19 & 45.75 \\
50\% DIQ & 62.45 & 54.08 & 73.28 & 63.27 & 14.56 & 51.30 & 47.73 & 33.67 & 46.03 & 46.61 & 39.98 & \textbf{47.75} \\ \midrule
10\% Influence                               & 64.81 & 54.84 & 75.94 & 65.20 & 14.56 & 49.35 & 41.23 & 15.06 & 44.82 & 49.75 & 35.80 & 45.60 \\
10\% Overall                           & 63.55 & 54.20 & 74.75 & 64.17 & 15.53 & 51.30 & 44.81 & 15.43 & 43.38 & 44.94 & 35.90 & 45.32 \\
10\% Knowledge                         & 63.71 & 53.91 & 75.67 & 64.43 & 18.41 & 47.40 & 46.75 & 17.10 & 44.51 & 48.36 & 37.09 & 46.20 \\
10\% Reasoning                         & 63.55 & 53.69 & 75.11 & 64.12 & 12.62 & 47.40 & 43.83 & 15.71 & 43.63 & 46.76 & 34.99 & 44.70 \\
10\% DIQ                        & 65.51 & 56.30 & 76.95 & 66.25 & 13.59 & 51.62 & 48.70 & 18.29 & 47.04 & 50.62 & 38.31 & \textbf{47.62} \\ \midrule
1\% Influence                                & 77.85 & 63.81 & 82.28 & 74.65 & 13.59 & 63.64 & 57.14 & 18.69 & 54.81 & 60.96 & 44.81 & 54.75 \\
1\% Overall                            & 77.53 & 64.04 & 81.63 & 74.40 & 14.56 & 64.61 & 55.19 & 18.78 & 54.74 & 61.76 & 44.94 & 54.76 \\
1\% Knowledge                          & 78.00 & 63.52 & 82.92 & 74.81 & 11.65 & 61.04 & 57.14 & 18.24 & 54.74 & 61.62 & 44.07 & 54.32 \\
1\% Reasoning                          & 76.28 & 63.69 & 83.10 & 74.36 & 10.68 & 64.94 & 57.47 & 17.71 & 55.13 & 61.18 & 44.52 & 54.46 \\
1\% DIQ                         & 77.45 & 63.93 & 82.74 & 74.71 & 15.53 & 66.88 & 58.12 & 18.53 & 54.74 & 61.76 & 45.93 & \textbf{55.52} \\ \bottomrule
\end{tabular}
\caption{Full downstream task results of Llama3.1-8B-Instruct and Qwen3-8B trained using datasets selected by different selection scores, under data keeping ratio of 1\%, 10\%, and 50\%.}
\label{tab:full_ablation}
\end{table*}

\subsection{Generalization Experiment}
\label{app:full_generalization}

Full downstream task results of cross-scale and cross-model experiment, and preference learning experiment are provided in Tables \ref{tab:full_cross} and \ref{tab:full_dpo}.

\begin{table*}[]
\centering
\small
\begin{tabular}{@{}lcccccccccccc@{}}
\toprule
\textbf{Model} &
  \multicolumn{1}{l}{\textbf{MedQ}} &
  \multicolumn{1}{l}{\textbf{MedM}} &
  \multicolumn{1}{l}{\textbf{MMLU}} &
  \multicolumn{1}{l}{\textbf{Avg$_S$}} &
  \multicolumn{1}{l}{\textbf{HLE}} &
  \multicolumn{1}{l}{\textbf{MeB4}} &
  \multicolumn{1}{l}{\textbf{MeB5}} &
  \multicolumn{1}{l}{\textbf{MedX}} &
  \multicolumn{1}{l}{\textbf{MedG}} &
  \multicolumn{1}{l}{\textbf{MetM}} &
  \multicolumn{1}{l}{\textbf{Avg$_C$}} &
  \multicolumn{1}{l}{\textbf{Avg$_A$}} \\ \midrule
\multicolumn{13}{c}{\textbf{Qwen3-8B}}                                                                             \\ \midrule
50\% Random        & 60.57 & 54.41 & 68.41 & 61.13 & 22.33 & 48.38 & 45.13 & 25.47 & 44.33 & 37.29 & 37.16 & 45.15 \\
50\% DIQ Llama Inf & 63.32 & 53.36 & 77.23 & 64.64 & 15.53 & 50.00 & 44.81 & 26.65 & 44.84 & 43.77 & 37.60 & 46.61 \\
50\% DIQ Qwen Inf  & 62.45 & 54.08 & 73.28 & 63.27 & 14.56 & 51.30 & 47.73 & 33.67 & 46.03 & 46.61 & 39.98 & 47.75 \\ \midrule
10\% Random        & 63.94 & 52.21 & 76.03 & 64.06 & 12.62 & 47.40 & 44.81 & 21.88 & 46.16 & 44.57 & 36.24 & 45.51 \\
10\% DIQ Llama Inf & 65.51 & 56.30 & 76.95 & 66.25 & 13.59 & 51.62 & 48.70 & 18.29 & 47.04 & 50.62 & 38.31 & 47.62 \\
10\% DIQ Qwen Inf  & 67.01 & 55.25 & 77.96 & 66.74 & 12.62 & 55.19 & 47.40 & 17.71 & 46.79 & 52.29 & 38.67 & 48.02 \\ \midrule
1\% Random         & 76.04 & 61.08 & 82.19 & 73.10 & 9.71  & 65.58 & 54.55 & 16.12 & 54.45 & 57.83 & 43.04 & 53.06 \\
1\% DIQ Llama Inf  & 76.28 & 63.93 & 82.74 & 74.32 & 13.59 & 66.88 & 57.47 & 17.71 & 54.74 & 60.96 & 45.23 & 54.92 \\
1\% DIQ Qwen Inf   & 76.67 & 64.45 & 83.38 & 74.83 & 15.53 & 64.29 & 57.14 & 18.12 & 54.96 & 61.18 & 45.20 & 55.08 \\ \midrule
\multicolumn{13}{c}{\textbf{Qwen3-14B}}                                                                            \\ \midrule
50\% Random        & 64.02 & 57.69 & 73.74 & 65.15 & 17.48 & 62.34 & 52.92 & 30.41 & 48.60 & 58.19 & 44.99 & 51.71 \\
50\% DIQ Llama Inf & 65.12 & 58.55 & 72.73 & 65.47 & 23.30 & 59.42 & 56.49 & 32.20 & 49.58 & 60.16 & 46.86 & 53.06 \\
50\% DIQ Qwen Inf  & 67.32 & 58.45 & 74.20 & 66.66 & 19.42 & 62.99 & 58.12 & 29.31 & 49.70 & 59.94 & 46.58 & 53.27 \\ \midrule
10\% Random        & 65.12 & 57.76 & 77.41 & 66.76 & 13.59 & 59.09 & 56.17 & 27.10 & 48.20 & 53.53 & 42.95 & 50.89 \\
10\% DIQ Llama Inf & 66.38 & 58.16 & 78.42 & 67.65 & 10.68 & 59.09 & 53.57 & 20.61 & 47.26 & 56.23 & 41.36 & 50.04 \\
10\% DIQ Qwen Inf  & 71.64 & 59.50 & 81.54 & 70.89 & 17.48 & 63.64 & 56.49 & 20.12 & 48.37 & 56.23 & 43.72 & 52.78 \\ \midrule
1\% Random         & 80.99 & 66.99 & 84.94 & 77.64 & 9.71  & 69.06 & 61.69 & 20.12 & 54.02 & 63.07 & 46.69 & 56.73 \\
1\% DIQ Llama Inf  & 81.38 & 67.32 & 85.95 & 78.22 & 6.80  & 69.06 & 61.69 & 20.20 & 54.90 & 63.15 & 45.97 & 56.72 \\
1\% DIQ Qwen Inf   & 82.09 & 67.73 & 85.31 & 78.38 & 10.68 & 73.38 & 64.29 & 20.61 & 55.25 & 64.02 & 47.42 & 58.15 \\ \midrule
\multicolumn{13}{c}{\textbf{Qwen3-32B}}                                                                            \\ \midrule
50\% Random        & 66.93 & 61.61 & 77.50 & 68.68 & 16.50 & 60.06 & 50.32 & 23.67 & 49.30 & 56.15 & 42.67 & 51.34 \\
50\% DIQ Llama Inf & 69.21 & 61.25 & 74.38 & 68.28 & 22.33 & 55.52 & 53.90 & 23.71 & 50.41 & 60.16 & 44.34 & 52.32 \\
50\% DIQ Qwen Inf  & 69.68 & 61.01 & 76.40 & 69.03 & 29.13 & 58.44 & 55.84 & 26.29 & 50.17 & 58.85 & 46.45 & 53.98 \\ \midrule
10\% Random        & 67.64 & 58.76 & 76.58 & 67.66 & 18.45 & 65.58 & 62.34 & 26.98 & 49.66 & 62.05 & 47.51 & 54.23 \\
10\% DIQ Llama Inf & 70.46 & 59.48 & 76.95 & 68.96 & 18.45 & 66.56 & 62.01 & 28.12 & 50.54 & 60.23 & 47.65 & 54.76 \\ 
10\% DIQ Qwen Inf  & 71.80 & 59.41 & 75.67 & 68.96 & 25.24 & 64.61 & 66.88 & 32.29 & 50.72 & 63.15 & 50.48 & 56.64 \\ \midrule
1\% Random         & 79.73 & 60.22 & 74.84 & 71.60 & 10.68 & 70.13 & 66.56 & 28.41 & 58.20 & 63.36 & 49.56 & 56.90 \\
1\% DIQ Llama Inf  & 78.38 & 59.05 & 78.42 & 71.95 & 13.59 & 68.83 & 64.29 & 24.08 & 57.43 & 64.31 & 48.76 & 56.49 \\
1\% DIQ Qwen Inf   & 78.24 & 61.25 & 81.54 & 73.68 & 16.50 & 66.88 & 64.61 & 26.57 & 57.33 & 63.22 & 49.19 & 57.35 \\ \bottomrule
\end{tabular}
\caption{Full downstream task results of Qwen series models trained using DIQ under different influence score settings.}
\label{tab:full_cross}
\end{table*}

\begin{table*}[]
\centering
\small
\begin{tabular}{@{}lllllllllllll@{}}
\toprule
\textbf{Model} &
  \textbf{MedQ} &
  \textbf{MedM} &
  \textbf{MMLU} &
  \textbf{Avg$_S$} &
  \textbf{HLE} &
  \textbf{MeB4} &
  \textbf{MeB5} &
  \textbf{MedX} &
  \textbf{MedG} &
  \textbf{MetM} &
  \textbf{Avg$_C$} &
  \textbf{Avg$_A$} \\ \midrule
FineMed           & 74.34 & 63.23 & 82.29 & 73.29 & 13.59 & 66.22 & 58.27 & 21.02 & 55.31 & 61.11 & 45.92 & 55.04 \\
FineMed + DPO     & 75.51 & 65.30 & 83.67 & 74.83 & 14.64 & 66.56 & 58.44 & 18.69 & 55.31 & 61.55 & 45.87 & 55.52 \\ \midrule
50\% Random + DPO & 78.16 & 63.71 & 82.74 & 74.87 & 9.71  & 66.56 & 58.44 & 18.61 & 55.31 & 61.11 & 44.96 & 54.93 \\
50\% DIQ + DPO    & 77.85 & 64.57 & 81.91 & 74.78 & 10.68 & 68.51 & 58.44 & 19.59 & 54.86 & 61.91 & 45.67 & 55.37 \\ \midrule
10\% Random + DPO & 76.75 & 63.88 & 83.47 & 74.70 & 8.74  & 65.26 & 60.39 & 18.69 & 54.67 & 62.86 & 45.10 & 54.97 \\
10\% DIQ + DPO    & 76.51 & 64.74 & 82.28 & 74.51 & 14.64 & 65.91 & 62.42 & 17.64 & 54.77 & 62.35 & 46.29 & 55.70 \\ \midrule
1\% Random + DPO  & 78.08 & 64.33 & 81.82 & 74.74 & 17.48 & 65.26 & 59.42 & 18.90 & 54.98 & 62.13 & 46.36 & 55.82 \\
1\% DIQ + DPO     & 76.98 & 64.57 & 84.02 & 75.19 & 16.50 & 65.58 & 64.29 & 18.90 & 54.96 & 62.86 & 47.18 & 56.52 \\ \bottomrule
\end{tabular}
\caption{Full downstream task resuls of Qwen3-8B fine-tuned using DIQ-selected or random data and DPO.}
\label{tab:full_dpo}
\end{table*} 


\end{document}